\documentclass[10pt,twocolumn,letterpaper]{article}

\usepackage[pagenumbers]{cvpr} % To force page numbers, e.g. for an arXiv version

\newcommand{\bff}{\mathbf{f}}\newcommand{\bF}{\mathbf{F}} %

\newcommand{\bH}{\mathbf{H}}

\newcommand{\bK}{\mathbf{K}}

\newcommand{\bm}{\mathbf{m}}

\newcommand{\bQ}{\mathbf{Q}}

\newcommand{\bV}{\mathbf{V}}
\newcommand{\bW}{\mathbf{W}}
\newcommand{\bX}{\mathbf{X}}

\newcommand{\boldparagraph}[1]{\vspace{0.1cm}\noindent{\bf #1:}}
\definecolor{cvprblue}{rgb}{0.21,0.49,0.74}
\usepackage[pagebackref,breaklinks,colorlinks,allcolors=cvprblue]{hyperref}
\usepackage{colortbl}
\usepackage{amsmath}
\usepackage{algorithm}
\usepackage{algpseudocode}
\usepackage{amsfonts}

\def\paperID{4778} % *** Enter the Paper ID here
\def\confName{CVPR}
\def\confYear{2025}

\title{FFAA: Multimodal Large Language Model based Explainable Open-World \\ Face Forgery Analysis Assistant}

\author{
Zhengchao Huang\textsuperscript{\rm 1}\quad
Bin Xia\textsuperscript{\rm 2}\quad
Zicheng Lin\textsuperscript{\rm 1}\quad
Zhun Mou\textsuperscript{\rm 1}\quad
Wenming Yang\textsuperscript{\rm 1}\thanks{The corresponding author}\quad
Jiaya Jia\textsuperscript{\rm 3}\\
\textsuperscript{\rm 1}Tsinghua University\qquad
\textsuperscript{\rm 2}The Chinese University of Hong Kong\quad
\textsuperscript{\rm 3}HKUST\\
\textbf{\small\url{https://ffaa-vl.github.io}}
}

\begin{document}
\maketitle
\begin{abstract}
The rapid advancement of deepfake technologies has sparked widespread public concern, particularly as face forgery poses a serious threat to public information security. However, the unknown and diverse forgery techniques, varied facial features and complex environmental factors pose significant challenges for face forgery analysis. Existing datasets lack descriptive annotations of these aspects, making it difficult for models to distinguish between real and forged faces using only visual information amid various confounding factors. In addition, existing methods fail to yield user-friendly and explainable results, hindering the understanding of the model's decision-making process. To address these challenges, we introduce a novel Open-World Face Forgery Analysis VQA (OW-FFA-VQA) task and its corresponding benchmark. To tackle this task, we first establish a dataset featuring a diverse collection of real and forged face images with essential descriptions and reliable forgery reasoning. Based on this dataset, we introduce FFAA: Face Forgery Analysis Assistant, consisting of a fine-tuned Multimodal Large Language Model (MLLM) and Multi-answer Intelligent Decision System (MIDS). By integrating hypothetical prompts with MIDS, the impact of fuzzy classification boundaries is effectively mitigated, enhancing model robustness. Extensive experiments demonstrate that our method not only provides user-friendly and explainable results but also significantly boosts accuracy and robustness compared to previous methods. 
\end{abstract}    
\section{Introduction}
\label{sec:intro}

The widespread availability of facial data online and access to deepfake technologies \cite{deepfake, faceswap} have facilitated the forgery of identities. Malicious actors employ deepfake technologies to manipulate faces for fraudulent purposes, generating false information and inciting public panic, which pose serious threats to both personal and public information security. To combat face forgery attacks, researchers have analyzed a variety of face forgery datasets \cite{ff++/xception,celeb,dfdc} and developed numerous advanced forgery detection models \cite{implicit,exposing,f3net,recce,dcl} utilizing deep learning techniques. 
\begin{figure}[t]
\centering
\includegraphics[width=.98\columnwidth]{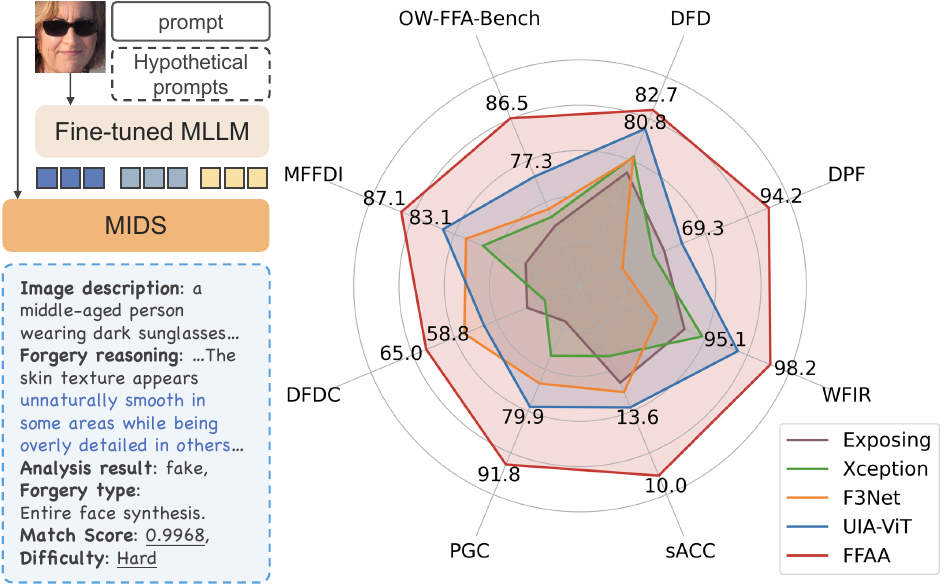}
\caption{Left: Architecture of \textbf{FFAA}. Right: FFAA achieves state-of-the-art generalization performance on OW-FFA-Bench (ACC=86.5\%) and exhibits excellent robustness (sACC=10.0\%).}
\label{fig:highlight}
\end{figure}

However, previous methods exhibit a substantial decline in detection accuracy when applied to open-world scenarios. There are two primary reasons: (1) High-quality forged faces crafted using unknown forgery techniques, demanding models with robust generalization. (2) Varied image quality, diverse facial attributes (\textit{e.g.}, makeup, accessories, expressions and orientation), and complex environmental factors necessitate models to exhibit robust anti-interference capabilities. In addition, previous methods typically treat it as a binary classification task. However, relying solely on binary classification results or heatmaps makes it difficult to understand the model's decision-making process.

Motivated by these, we introduce a novel \textbf{O}pen-\textbf{W}orld \textbf{F}ace \textbf{F}orgery \textbf{A}nalysis \textbf{VQA} (OW-FFA-VQA) task, extending the original binary classification task into a VQA task that, given any face image, requires authenticity determination and an analysis process with supporting evidence. To support this task, we employ GPT4-assisted data generation to create the \textbf{FFA-VQA}, an explainable open-world face forgery analysis VQA dataset, which includes a diverse collection of real and forged face images with essential descriptions and reliable forgery reasoning. Based on this dataset, we introduce \textbf{FFAA}: \textbf{F}ace \textbf{F}orgery \textbf{A}nalysis \textbf{A}ssistant. Specifically, to endow the model with comprehensive face forgery analysis capabilities and enable flexible responses based on varying hypotheses, we fine-tune an MLLM on the FFA-VQA dataset containing hypothetical prompts. Subsequently, \textbf{M}ulti-answer \textbf{I}ntelligent \textbf{D}ecision \textbf{S}ystem (MIDS) is proposed to select the answer that best matches the image's authenticity from MLLM's responses under different hypotheses, mitigating the impact of fuzzy classification boundaries between real and forged faces, thereby enhancing model robustness.

To evaluate model performance in complex real-world conditions, we establish a challenging \textbf{O}pen-\textbf{W}orld \textbf{F}ace \textbf{F}orgery \textbf{A}nalysis benchmark (OW-FFA-Bench), comprising a diverse set of real and forged face images from seven public datasets. Ablation studies are conducted to validate the effectiveness of our dataset and MIDS's efficacy. Furthermore, we assess the interpretability of our method by comparing it with advanced MLLMs and visualizing attention heatmaps. Examples are also provided to demonstrate FFAA’s capability to distinguish between easy and challenging classification samples without threshold adjustments. Our method not only provides user-friendly, explainable results but also achieves notable improvements in accuracy and robustness over previous methods. To our knowledge, we are the first to explore and effectively utilize fine-tuned MLLMs for explainable face forgery analysis. In summary, our contributions are as follows:

\begin{itemize}
\item We introduce a novel OW-FFA-VQA task and establish the OW-FFA-Bench for evaluation. This task is essential for understanding the model's decision-making process and advancing real-world face forgery analysis.
\item We utilize GPT4-assisted data generation to create the FFA-VQA dataset, featuring diverse real and forged face images with essential descriptions and forgery reasoning. 
\item We propose FFAA, consisting of a fine-tuned MLLM and MIDS. By integrating hypothetical prompts with MIDS, the impact of fuzzy classification boundaries is effectively mitigated, enhancing robustness.
\item Extensive experiments demonstrate that our method significantly enhances accuracy and robustness while providing explainable results compared to previous methods.
\end{itemize}

\section{Related Work}
\label{sec:related_work}

\boldparagraph{Face Forgery Analysis Datasets and Methods} FF++ \cite{ff++/xception} is the first large-scale face forgery video dataset, which includes four types of forgery techniques: DeepFakes \cite{deepfake}, FaceSwap \cite{faceswap}, Face2Face \cite{face2face} and NeuralTextures \cite{nt}. Afterwards, some real-world datasets emerge, such as DFDC \cite{dfdc}, Deeperforensics \cite{deeperforensics}. Although these datasets providing numerous images or videos, they lack the necessary textual descriptions. Recently, \citeauthor{ddvqa} construct the DD-VQA dataset by manually incorporating reasoning process for certain real and forged faces discernible by common sense. However, manual annotation is quite cumbersome and prone to bias. Additionally, numerous face forgeries in open-world scenarios are challenging to identify using common sense alone and require expert scrutiny. 

Despite the datasets, researchers have previously proposed numerous methods \cite{xception, multiattn, recce, twostream, dcl} for detecting face forgeries. However, these methods exhibit limited generalization in open-world scenarios. To improve generalization, researchers have made improvements from various perspectives. \citeauthor{implicit} find that inadvertently learned identity representations in images hinder generalization thus design a specialized network to mitigate this impact. \citeauthor{fg} leverages annotated demographic information \cite{fg-dataset} and employs disentanglement learning to extract domain-agnostic forgery features and demographic features, fostering fair learning. However, these methods still demonstrate limited generalization in complex open-world scenarios. Additionally, these methods typically approach the problem as a binary classification task, yielding solely binary outputs or heatmaps, making it difficult to understand the model's decision-making process. Therefore, it is essential to develop an open-world face forgery analysis model with strong generalization and robustness, while also providing user-friendly and explainable results.

\boldparagraph{Multimodal Large Language Models} Recently, Multimodal Large Language Models (MLLMs) represented by GPT-4V \cite{gpt4v} and GPT-4o \cite{gpt4o} have garnered significant attention due to their remarkable image understanding and analysis capabilities. Some studies \cite{cangpt,shield} have explored their potential in face forgery analysis by employing prompt engineering, finding that well-crafted prompts are crucial for analysis and judgment, though discrimination accuracy remains suboptimal. In other fields, some studies \cite{lisa, llava-med} employ instruction tuning to fine-tune pre-trained MLLMs, resulting in domain-specific MLLMs with strong zero-shot capabilities. However, the fine-tuning of MLLMs for face forgery analysis and the strategies to leverage their advantages remain underexplored.
\section{Dataset}
\label{sec:dataset}

\begin{figure*}[t]
\centering
\includegraphics[width=\textwidth]{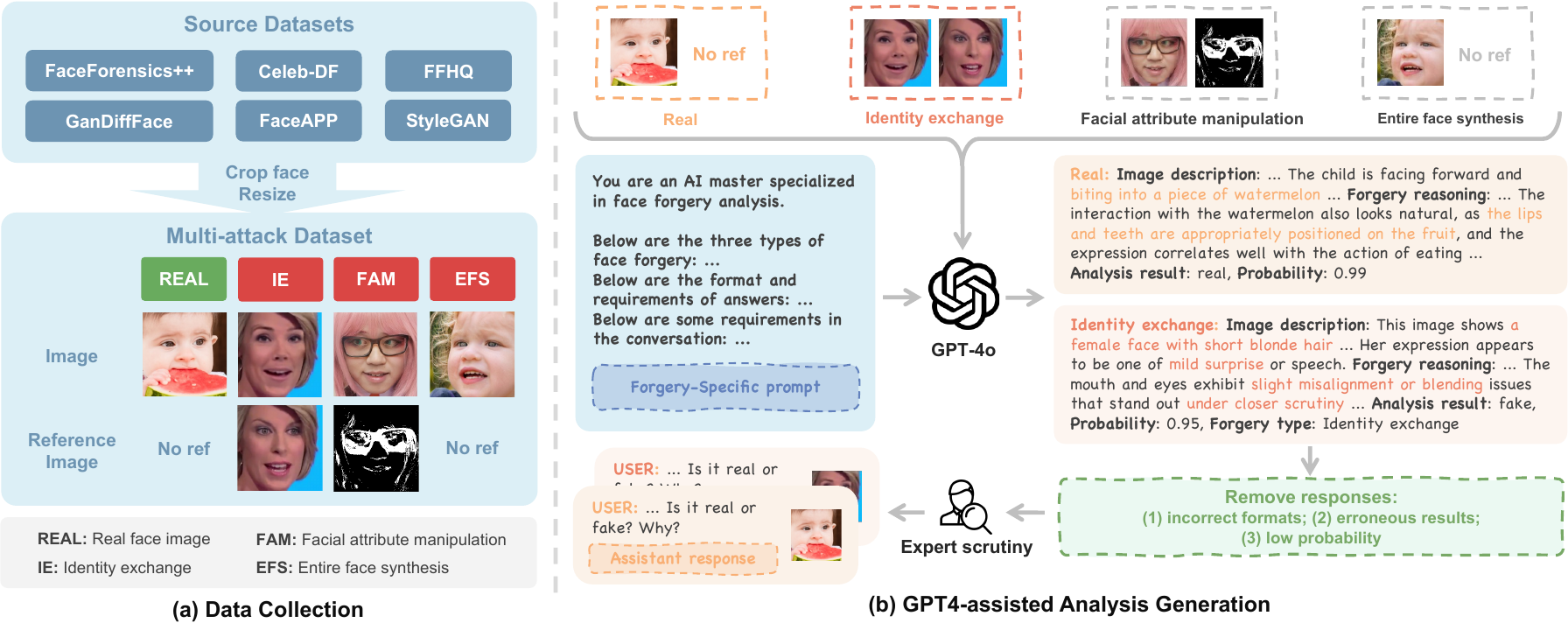}
\caption{Construction pipeline for the Multi-attack dataset (Left) and the FFA-VQA dataset (Right).}
\label{dataset_pipeline}
\end{figure*}

\subsection{Data Collection}
Fig.~\ref{dataset_pipeline}a shows the data collection process. Based on some studies \cite{ffdsurvey, dffd} and real-world scenarios, we classify face forgery types into three categories: identity exchange (\textit{i.e.}, replacing the target face with another face), facial attribute manipulation (\textit{i.e.}, manipulating attributes of the target face, including expressions), and entire face synthesis (\textit{i.e.}, generating non-existent faces). Then, we utilize FF++ \cite{ff++/xception}, Celeb-DF-v2 \cite{celeb}, DFFD \cite{dffd} and GanDiffFace \cite{gandiffface} as our source datasets since they encompass diverse forgery techniques and enriched facial characteristics. Finally, we create the \textbf{M}ulti-\textbf{a}ttack (MA) dataset, consisting of 95K images with diverse facial features and multiple forgery types.

\subsection{GPT4-assisted Analysis Generation}
Manually writing the image description and forgery reasoning requires specialized knowledge and is quite cumbersome, making it difficult to establish a large-scale and high-quality dataset. Inspired by some studies \cite{sharegpt4v, llava, llava-med} using GPT for image captioning, we employ GPT-4o \cite{gpt4o} to generate face forgery analysis. The pipeline of analysis generation is shown in Fig.~\ref{dataset_pipeline}b, which can be summarized as follows. First, a system prompt, forgery-specific prompt, target image and available reference images are integrated as inputs to GPT-4o to produce responses in a structured format. Responses with incorrect formats, erroneous results, or low probability are filtered out. Finally, experts scrutinize the analysis, and qualified responses are organized into a VQA format. \textit{See Supplementary Material for more details}.

\boldparagraph{System Prompt Design} Given the challenges of face forgery analysis in open-world scenarios, it is crucial to supply domain-specific auxiliary information and carefully crafted prompts to guide the generation. Motivated by these, our system prompt comprises four sections: role-playing, the introduction to forgery types, the Chain-of-Thought and conversation requirements. To leverage relevant domain knowledge of GPT-4o and generate responses with domain-specific characteristics, we first introduce its role and the task. Then, the types of face forgeries are defined to avoid ambiguity and assist in the analysis. For the Chain-of-Thought, we advocate that conducting complex face forgery analysis demands a robust foundation in image understanding, requiring multi-perspective analysis before delivering the final discrimination result. Thus, we design a progressive \textbf{F}ace \textbf{F}orgery \textbf{A}nalysis \textbf{C}hain \textbf{o}f \textbf{T}hought (FFA-CoT) consisting of \textit{'Image description'}, \textit{'Forgery reasoning'} and \textit{'Analysis result'} (including classification result, probability and forgery type). Finally, some requirements are claimed to ensure the consistency of the responses.

\begin{figure}[t]
\centering
\includegraphics[width=1.0\columnwidth]{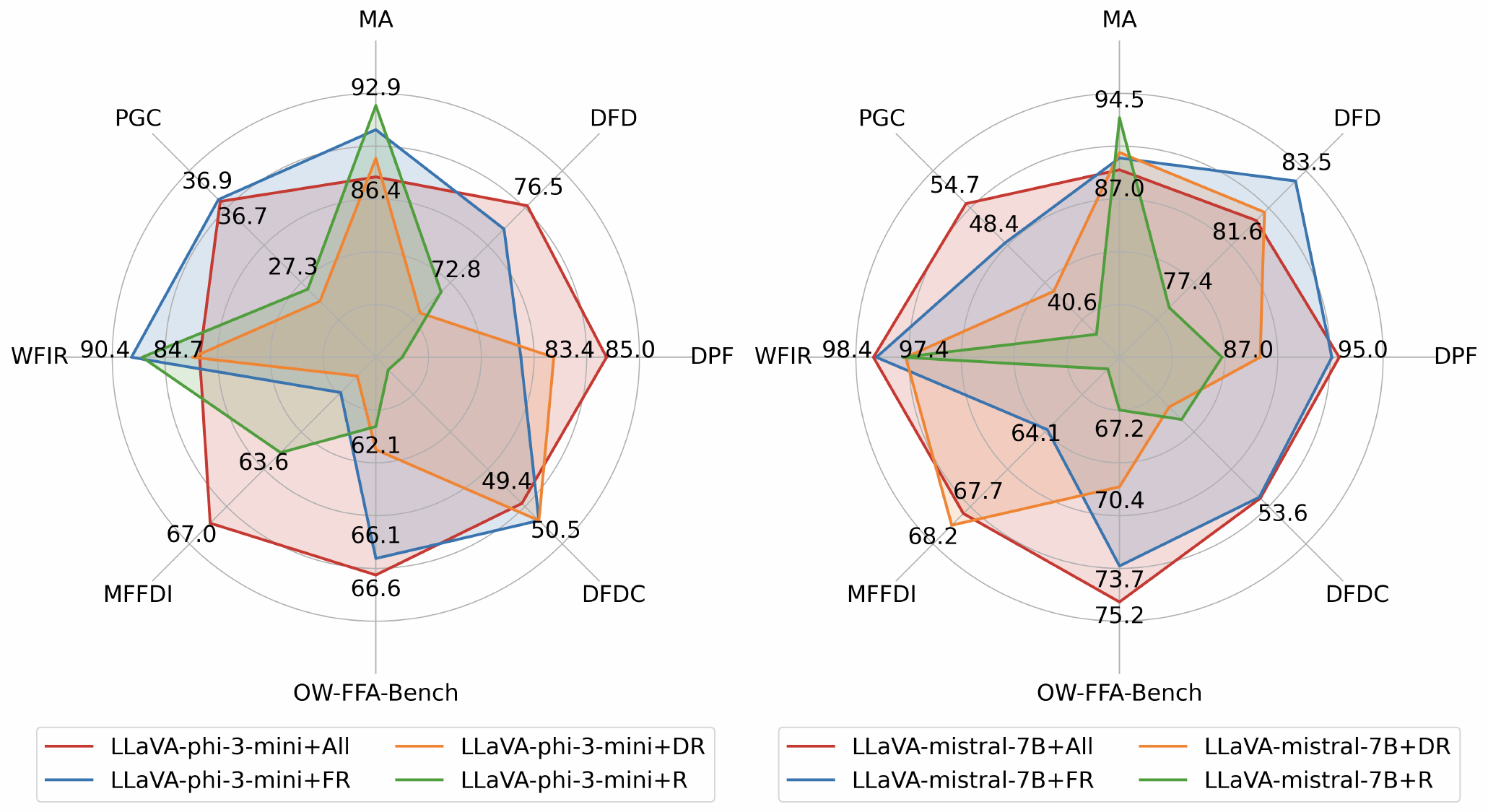}
\caption{FFA-VQA endows MLLMs with powerful face forgery analysis capabilities in open-world scenarios.}
\label{dataset_quality}
\end{figure}

\begin{figure*}[t]
\centering
\includegraphics[width=\textwidth]{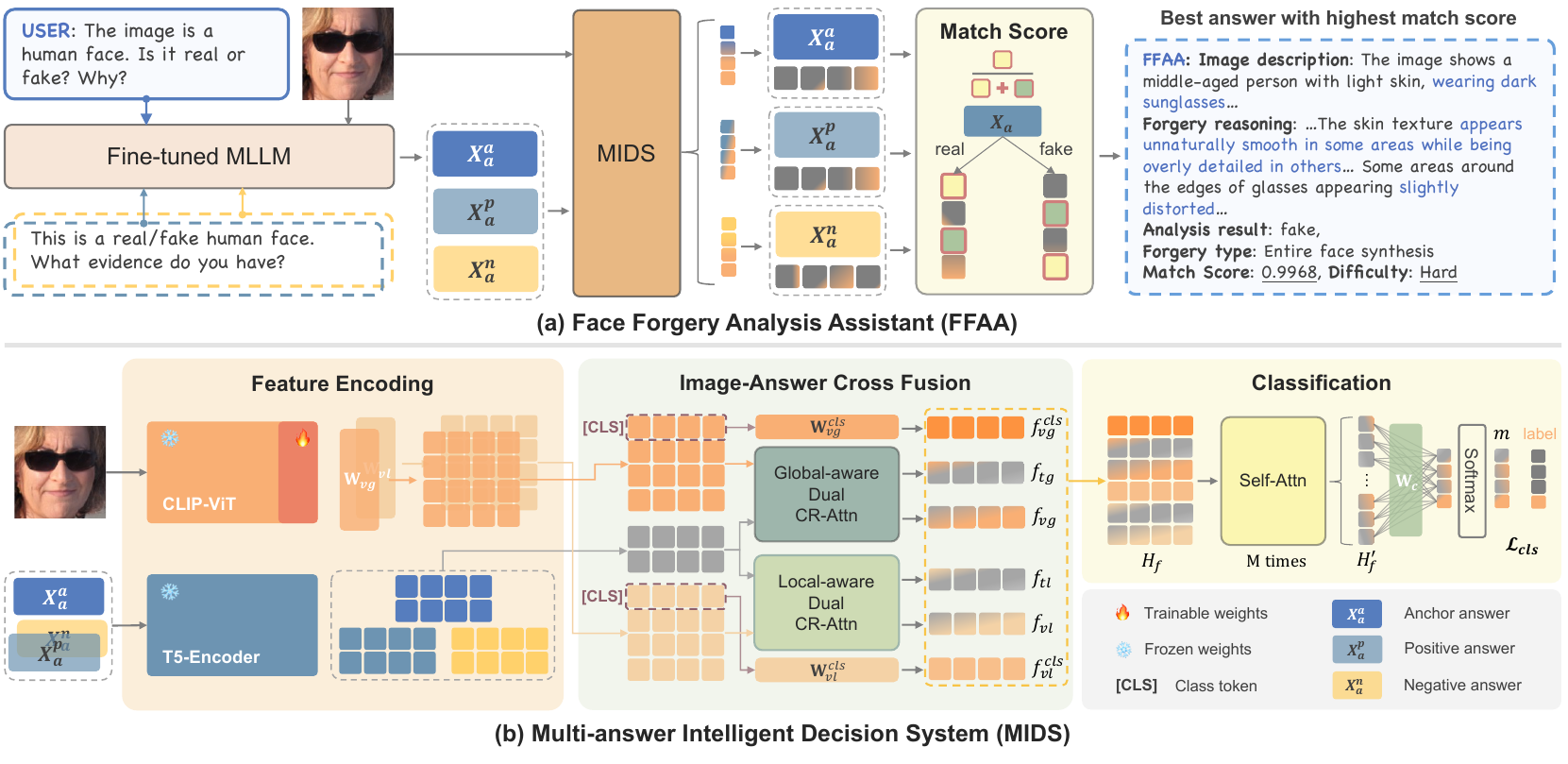}
\caption{The workflow of FFAA (Top) and the architecture of MIDS (Bottom).}
\label{fig:method}
\end{figure*}

\boldparagraph{Bias Mitigation Strategies} (1) \textit{Annotator bias}: We use GPT-4o with a zero-shot in-context learning paradigm for annotation, mitigating annotator subjective bias; (2) \textit{Generated analysis bias}: To enhance response reliability, we include the authenticity and forgery type of the target image as prior information in the forgery-specific prompt (\textit{e.g.}, \textit{'The image is a fake face whose forgery type is identity exchange. Please analyze...'}), along with the original image or forged mask as a reference for more precise analysis. (3) \textit{Analysis result bias}: Responses not conforming to FFA-CoT are filtered to ensure structural consistency. Then responses with erroneous results are removed. Lastly, lower probability often indicate higher analysis difficulty and reduced reliability. Thus, responses with probability values below a set threshold are excluded; (4) \textit{Analysis rationality bias}: Experts scrutinize responses with access to the original image, its authenticity and forgery type, and reference images to assess the rationality of the forgery analysis. Responses deemed unreasonable by experts are filtered out to mitigate potential biases in GPT-4o's analysis.

Finally, we obtain 20K high-quality face forgery analysis data which are then supplemented with semantically identical inquiry prompts to form the \textbf{FFA-VQA} dataset.

\boldparagraph{Quality Verification} We conduct experiments to verify the quality of FFA-VQA. As depicted in Fig.~\ref{dataset_quality}, incorporating high-quality image descriptions and forgery reasoning enables MLLMs to perform face forgery analysis based on image understanding before making final judgments, which effectively reduces overfitting and enhances generalization. \textit{See Sec.\ref{subsec: ab-ffa-vqa} for more experimental details}.

\section{Method}
\label{sec:method}

Although fine-tuning MLLMs on FFA-VQA effectively enhances generalization, model robustness remains suboptimal. We also observe that prompts with similar semantics but varied content can yield different authenticity judgments for faces that are challenging to discern. This variability arises from the advancement of forgery techniques and the complexities of open-world scenarios, which have considerably blurred the classification boundaries between real and forged faces. Motivated by these, we introduce hypothetical prompts into FFA-VQA, presuming the face to be either real or fake prior to analysis. Fine-tuning the MLLM on this dataset enables answer generation based on different hypotheses. Subsequently, we propose \textbf{MIDS} to select the answer that best aligns the face's authenticity, which effectively mitigates the impact of fuzzy classification boundaries between real and forged faces. Fig.~\ref{fig:method}a depicts the architecture of \textbf{FFAA}, which primarily consists of two modules: a fine-tuned MLLM and MIDS. Initially, the MLLM generates the anchor answer $\bX_a^a$ without hypotheses, the positive answer $\bX_a^p$ and the negative answer ${\bX_a^p}$ based on varying hypotheses, all of which are input into MIDS along with the face image. The model then calculates the match score $S_m$ between the image and each answer based on MIDS's output vectors $(\bm^a, \bm^p, \bm^n)$ and the determination result of each answer as follows:
\begin{equation}
\label{eq:match-score}
S_m =
\begin{cases} 
    m_0 / (m_0 + m_2), & \text{if}~~ \mathcal{R}(\bX_a)~~\text{is}~~'real', \\
    m_3 / (m_3 + m_1), & \text{if}~~ \mathcal{R}(\bX_a)~~\text{is}~~'fake',
\end{cases}
\end{equation}
where $m_i$ represents the $i$-th element of $\bm$ and $\mathcal{R}(\bX_a)$ represents the determination result of $\bX_a$. Finally, the answer with the highest match score is selected as the best answer.

\subsection{Fine-tuning MLLM with Hypothetical Prompts}
We categorize the prompts into two types: non-hypothetical (\textit{e.g.}, \textit{The image is a human face image. Is it real or fake? Why?}) and hypothetical (\textit{e.g.}, \textit{This is a [real/fake] human face. What evidence do you have?}). One-third of the conversations from FFA-VQA are randomly selected, with their prompts modified from non-hypothetical to hypothetical, aligning the hypothesis with the authenticity of the face. Then, we fine-tune an MLLM on this dataset with LoRA \cite{lora}. For the image $\bX_v$, its conversation data is $(\tilde{\bX}_q, \bX_a)$, where $\tilde{\bX}_q=\bX_q$ for non-hypothetical prompts and $\tilde{\bX}_q=[\bX_h, \bX_q]$ for hypothetical prompts. The fine-tuning objective is consistent with the original auto-regressive training objective of the LLM. For a sequence of length $L$, the probability of the target answer is computed as:
\begin{equation}
\label{eq:lora}
    p(\bX_a|\bX_v, \tilde{\bX}_q) = \prod_{i=1}^L p_{\boldsymbol{\theta}}(x_i|\bX_v, \tilde{\bX}_q, \bX_{a,<i}),
\end{equation}
where $\boldsymbol{\theta}$ is the trainable parameters and $\bX_{a,<i}$ are answer tokens before current prediction token $x_i$. From the conditionals in Eq.~\ref{eq:lora}, it is evident that for prompts with hypothesis $\bX_h$, $\bX_a$ depends on both $\bX_h$ and $\bX_v$. For faces that are challenging to discern, where altering $\bX_h$ may result in varying authenticity judgments, we label them as \textit{'Hard'}. For easily distinguishable faces, even if altered, the model is likely to rely on the high-weight image tokens in $\bX_v$ to maintain its original judgment. We label these as \textit{'Easy'}. Therefore, by simply altering $\bX_h$, we can identify faces that are challenging to distinguish and subsequently make a more informed decision among multiple answers.

\subsection{Multi-answer Intelligent Decision System}
MIDS consists of three components: Feature Encoding, Image-Answer Cross Fusion Module and Classification Layer, as depicted in Fig.~\ref{fig:method}b. We first utilize the fine-tuned MLLM to extract answers from unused samples in the Multi-attack dataset for training. Specifically, for a given image $\bX_v$, a non-hypothetical prompt is randomly selected to query the MLLM, yielding $\bX_a^a$. Based on the determination in $\bX_a^a$, two hypothetical prompts assuming the same or opposite determination are used to query again, yielding the positive answer $\bX_a^p$ and the negative answer $\bX_a^{n}$ respectively. 
Finally, based on the authenticity $y_v\in\{0,1\}$ of $\bX_v$ and the determination result $y_a\in\{0,1\}$ of $\bX_a$, we assign a label $y$ for each $(\bX_v, \bX_a)$ as:
\begin{equation}
\label{eq:mids-label}
y = 2y_v+y_a\in\{0,1,2,3\},
\end{equation}
\boldparagraph{Feature Encoding} For each given answer $\bX_a$, to prevent the model from relying solely on the final classification result while ignoring the intermediate analysis process, we mask the \textit{'Analysis result'} of $\bX_a$. Then, we employ the pre-trained T5-Encoder \cite{t5} with frozen weights for good linguistic consistency to obtain the text feature $\bF_t$. For the image $\bX_v$, we employ the pre-trained CLIP-ViT-L/14 \cite{clip} as the visual feature encoder. To maintain basic image understanding and enhance the perception of local and global features related to face authenticity, the weights of the last two layers are unfrozen. Then, trainable projection matrices $(\bW_{vg}, \bW_{vl})$ are employed to project the output features of the last two layers into the same dimensional space as $\bF_t$, resulting in $\bF_{vl}$ and $\bF_{vg}$.

\boldparagraph{Image-Answer Cross Fusion} To extract the corresponding visual features based on the answer and to identify matching textual features from the visual features, we employ dual cross-attention for comprehensive fusion across all dimensions. 
Specifically, it can be expressed as:
\begin{equation}
\label{eq:cr-attn}
\begin{aligned}
&\bF_{v\to t} = \text{LayerNorm}(\text{softmax}(\frac{\bQ_{v}\bK_t^T}{\sqrt{d}})\bV_t), \\
&\bF_{t\to v} = \text{LayerNorm}(\text{softmax}(\frac{\bQ_{t}\bK_v^T}{\sqrt{d}})\bV_v), \\
\end{aligned}
\end{equation}
where $d$ is the dimension of these features, $\{\bQ_{t}, \bK_{t}, \bV_{t}\}=\mathcal{F}_{qkv}^{t}(\bF_{t})$, $\{\bQ_{v}, \bK_{v}, \bV_{v}\}=\mathcal{F}_{qkv}^{v}(\bF_{v})$. We perform the operations of Eq.~\ref{eq:cr-attn} with $(\bF_t, \bF_{vl})$ and $(\bF_t, \bF_{vg})$ respectively, to obtain $\bF_{t\to vl}$, $\bF_{vl\to t}$, $\bF_{t\to vg}$ and $\bF_{vg\to t}$.  Subsequently, we employ learnable vectors $\bff^0$ to obtain the corresponding aggregated feature based on cross-attention:
\begin{equation}
\label{eq:avg-features}
\begin{aligned}
&\bff^{r}=\text{LayerNorm}(\text{softmax}(\frac{\bQ_{f}\bK_{F}^T}{\sqrt{d}})\bV_{F}), \\
&\bQ_f=\bW_f^Q\bff^{r-1},~
\bK_F=\bW_F^K\bF,~
\bV_F=\bW_F^V\bF,~ \\
\end{aligned}
\end{equation}
where $r$ denotes the $r$-th iteration. Through the operations of Eq.~\ref{eq:avg-features}, we obtain text feature-enhanced visual domain features $(\bff_{t\to vl}^r, \bff_{t\to vg}^r)$ and image feature-enhanced textual domain features $(\bff_{v\to tl}^r, \bff_{v\to tg}^r)$ for each $(\bX_v, \bX_a)$.

\boldparagraph{Classification and Loss Function} To select the best answer, the model must possess a certain ability to discern the authenticity of faces and answers. Therefore, we set the classification labels as the four-class categorization shown in Eq.~\ref{eq:mids-label}. The [CLS] tokens of $\bF_{vl}$ and $\bF_{vg}$, containing enriched image category information, are passed through projection matrices $(\bW_{vl}^{cls}, \bW_{vg}^{cls})$ to obtain $(\bff_{vl}^{cls}, \bff_{vg}^{cls})$. Then, we concatenate $(\bff_{vg}^{cls}, \bff_{t\to vg}^r, \bff_{v\to tg}^r, \bff_{vl}^{cls}, \bff_{t\to vl}^r, \bff_{v\to tl}^r)$ to obtain $\bH_f$, which are further enhanced by $M$ iterations of self-attention with residual connections \cite{resnet}, resulting in $\bH'_f$.
Finally, $\bH'_f$ is passed through $\bW_c$ and a softmax to obtain $\bm\in \mathbb{R}^{4}$. The loss is computed as follows:
\begin{equation}
\label{eq:cls-loss}
\mathcal{L}_{cls}=-\frac{1}{N}\sum_{i=1}^{N}\sum_{c}^4 y_{i,c}\log(m_{i,c}),
\end{equation}
where $N$ is the number of samples, $y_{i,c}$ is the true label for the $i$-th sample in the $c$-th class and $m_{i,c}$ is the predicted probability for the $i$-th sample in the $c$-th class.

\renewcommand{\arraystretch}{1.1}
\begin{table*}[t]
\centering
\resizebox{\textwidth}{!}{
\begin{tabular}{ccccccccccc}
\toprule
\textbf{Category} & \textbf{Method} & \textbf{DPF} & \textbf{DFD} & \textbf{DFDC} & \textbf{PGC} & \textbf{WFIR} & \textbf{MFFDI} & \textbf{ALL} & \textbf{MA} & \textbf{sACC}$\downarrow$ \\
\midrule
Naive & ResNet-34\cite{resnet} & 33.8 / 30.2 & 74.6 / 80.3 & 58.0 / 69.6 & 36.1 / 55.4 & 78.1 / 92.2 & 64.1 / 70.8 & 57.4 / 61.1 & 95.6 / 99.2 & 20.8 \\
Naive & EfficientNet-B4\cite{efficientnet} & 30.3 / 24.8 & 79.8 / 87.2 & 51.5 / 75.0 & 62.5 / 94.9 & 96.5 / \underline{99.5} & 78.0 / 86.7 & 66.4 / 75.2 & \textbf{98.8} / \textbf{99.9} & 22.9 \\
Naive & Xception\cite{xception} & 61.1 / 60.2 & 78.0 / 87.8 & 45.7 / 74.8 & 69.4 / 96.5 & 91.7 / 98.5 & 79.3 / 86.5 & 70.9 / 75.2 & 96.8 / 99.4 & 16.3 \\
\midrule
Spatial & RECCE\cite{recce}  & 39.9 / 28.7 & 72.2 / 82.8 & 41.5 / 73.7 & \textbf{93.6} / 98.0 & 91.6 / 97.4 & 75.9 / 83.6 & 69.1 / 74.0 & 98.1 / 99.6 & 22.4 \\
Spatial & DCL\cite{dcl}  & 57.9 / 63.9 & \underline{82.5} / \underline{89.3} & 47.5 / 78.5 & 41.2 / 99.2 & \underline{98.0} / \textbf{99.8} & 78.1 / 89.8 & 67.5 / 76.2 & 98.2 / 99.6 & 21.6 \\
Spatial & Implicit*\cite{implicit} & 46.7 / 46.5 & 77.5 / 87.7 & 53.7 / 70.8 & 56.6 / 67.0 & 49.6 / 51.1 & 67.8 / 78.1 & 58.7 / 63.3 & 67.8 / 70.2 & \underline{10.4} \\
Spatial & FedForgery*\cite{fedforgery} & 75.8 / 83.5 & 76.6 / 88.1 & 38.6 / 58.2 & 68.5 / 55.9 & 48.9 / 47.8 & 52.5 / 56.8 & 60.1 / 67.8 & 51.0 / 64.6 & 13.7 \\
Spatial & Exposing\cite{exposing} & 64.2 / 75.5 & 76.4 / 86.2 & 48.6 / 77.2 & 62.3 / 94.8 & 90.0 / 98.0 & 75.2 / 83.4 & 69.5 / 76.8 & 94.5 / 98.8 & 14.9 \\
Spatial & NPR\cite{npr} & 47.5 / 44.2 & 67.2 / 74.5 & 40.1 / 66.4 & 67.2 / 80.1 & 91.6 / 99.3 & 71.4 / 79.4 & 64.2 / 72.5 & 97.1 / 99.7 & 19.3 \\
Spatial & UIA-ViT\cite{uiavit} & 69.3 / 74.6 & 80.8 / 88.5 & 55.6 / \textbf{87.6} & 79.9 / \underline{99.4} & 95.1 / 99.1 & \underline{83.1} / 90.7 & \underline{77.3} / \underline{84.4} & \underline{98.4} / \underline{99.8} & 13.6 \\
Spatial & CLIPping\cite{clipping} & \underline{90.4} / \underline{97.6} & 77.8 / 86.6 & 64.0 / 71.9 & 46.0 / 97.1 & 91.5 / 98.0 & 55.6 / \underline{96.4} & 70.9 / 80.9 & 87.4 / 94.2 & 16.9 \\
\midrule
Frequency & F3Net\cite{f3net} & 52.2 / 52.7 & 77.9 / 84.6 & 58.8 / \underline{79.9} & 75.1 / 96.5 & 87.4 / 95.1 & 80.9 / 89.0 & 72.1 / 80.1 & 96.8 / 99.5 & 14.4 \\
Frequency & Two-Stream\cite{twostream} & 50.8 / 57.7 & 75.0 / 80.6 & 47.1 / 72.1 & 52.7 / 87.9 & 86.4 / 96.4 & 77.8 / 85.5 & 65.0 / 72.5 & 97.5 / 99.6 & 18.1 \\
Frequency & FADD\cite{fadd} & 59.0 / 61.6 & 67.6 / 69.6 & \textbf{76.3} / 59.4 & 14.8 / 20.0 & 86.3 / 97.7 & 62.5 / 70.7 & 61.1 / 68.9 & 89.6 / 94.0 & 23.2 \\
\midrule
Ours & FFAA (w/o MIDS) & \cellcolor{gray!30}95.0 / 96.0 & \cellcolor{gray!30}81.6 / 83.9 & \cellcolor{gray!30}53.6 / 72.4 & \cellcolor{gray!30}54.7 / 75.2 & \cellcolor{gray!30}98.4 / 98.8 & \cellcolor{gray!30}67.7 / 73.7 & \cellcolor{gray!30}75.2 / 77.8 & \cellcolor{gray!30}87.0 / 88.7 & \cellcolor{gray!30}17.0\\
Ours & FFAA & \textbf{94.2} / \textbf{98.8} & \textbf{82.7} / \textbf{92.0} & \underline{65.0} / 74.0 & \underline{91.8} / \textbf{99.7} & \textbf{98.2} / \textbf{99.8} & \textbf{87.1} / \textbf{98.6} & \textbf{86.5} / \textbf{94.4} & 86.8 / 96.7 & \textbf{10.0}\\
\bottomrule
\end{tabular}
}
\caption{\small Comparison results (ACC / AUC) (\%) on OW-FFA-Bench and the test set of MA. * denotes the use of trained models provided by the authors, while the others were trained on MA. Our method was trained on FFA-VQA and (image, answers) data from the fine-tuned MLLM, with all images sourced from MA. Comparing our final method with other methods, bold and underline highlight the best and second-best performances, respectively. The gray-shaded areas indicate the performance of FFAA without MIDS.}
\label{baseline}
\end{table*}

\section{Experiments}
\label{sec:experiments}

In this section, we first establish \textbf{OW-FFA-Bench} to evaluate model generalization and robustness. We then compare our method with the state-of-the-art methods on this benchmark and conduct ablation studies. Furthermore, we perform qualitative studies by comparing FFAA with advanced MLLMs and employing several visualizations.

\begin{table}[t]
\centering
\resizebox{\columnwidth}{!}{
\begin{tabular}{cccc}
\toprule
\textbf{Name}       & \textbf{DPF}                 & \textbf{DFD}               & \textbf{DFDC}      \\
\midrule
Real: Fake & 466:534             & 327:673           & 198:802   \\
Source     & Deeperforensics\cite{deeperforensics} & DeepFakeDetection\cite{deepfakedetection} & DFDC\cite{dfdc}      \\
\midrule
\textbf{Name}       & \textbf{PGC}                 & \textbf{WFIR}              & \textbf{MFFDI}     \\
\midrule
Real: Fake & 759:241             & 499:501           & 485:515   \\
Source     & PGGAN\cite{pggan}, Celeb-A\cite{celeba}      & WhichFaceIsReal\cite{whichfaceisreal}   & MultiFFDI\cite{multiffdi} \\
\bottomrule
\end{tabular}
}
%}
\caption{\small The data distribution and source of OW-FFA-Bench.}
\label{benchmark}
\end{table}

\subsection{Experimental Setup}

\boldparagraph{Datasets} We create the generalization test sets by randomly collecting face images from seven public datasets. The images are then divided into six test sets based on their source as shown in Tab.~\ref{benchmark}, with each containing 1K images and differing in distribution.

\boldparagraph{Implementation details}
We first fine-tune an MLLM, \textit{e.g.} LLaVA-v1.6-mistral-7B \cite{llava1.5} with LoRA (rank=32, alpha=48) on FFA-VQA using 2 RTX 3090 GPUs for 3 epochs with a learning rate of 1e-4 and a batch size of 16. Then, we train MIDS using 2 RTX 3090 GPUs for 2 epochs with a learning rate of 1e-4 and a batch size of 48. \textit{See Supplementary Material for more training details}.

\boldparagraph{Evaluation metrics}
We utilize Accuracy (ACC) and Area Under Receiver Operating Characteristic Curve (AUC) to empirically evaluate generalization performance on individual test sets and across the entire OW-FFA-Bench (ALL). Additionally, we employ the standard deviation of ACC (sACC) across all test sets to assess model robustness.

\begin{figure*}[t]
\centering
\includegraphics[width=\textwidth]{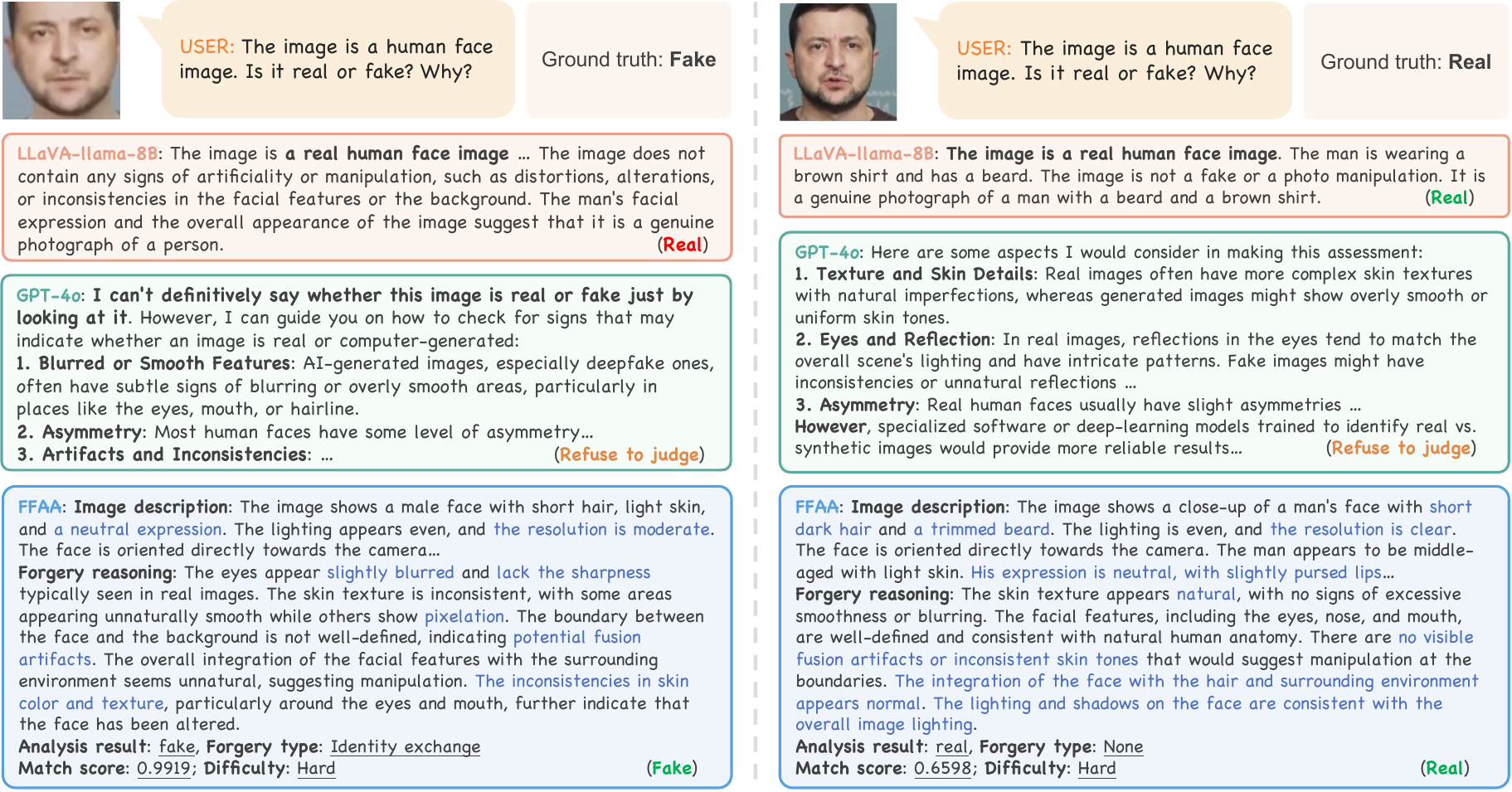}
\caption{\small Qualitative examples in real-world scenarios. Left: a facial frame from a video with a deepfake face of Zelenskyy delivering false statements. Right: a facial frame from a video of Zelenskyy’s genuine speech. Judgment results are indicated in parentheses as (\textit{'Real'}, \textit{'Fake'}, \textit{'Refuse to judge'}), with green for correct judgments and red for incorrect ones.}
\label{fig:answer_visual}
\end{figure*}

\subsection{Comparison with Competitive Methods}
In this section, we compare our method against state-of-the-art face forgery analysis methods on 1K in-domain test set of MA and OW-FFA-Bench. These methods fall into three categories: (1) \textit{Naive}: utilizing conventional classification models for binary classification; (2) \textit{Spatial}: incorporating specialized modules or additional constraints to facilitate the extraction of forgery cues at the spatial level, followed by classification; (3) \textit{Frequency}: leveraging the frequency domain to identify forgery characteristics, which are then employed for classification.

Tab.~\ref{baseline} presents the comparison results. These three categories of methods exhibit varying degrees of overfitting and limited generalization (\textit{i.e.}, strong performance on in-domain MA yet significantly reduced on cross-domain OW-FFA-Bench). Additionally, substantial variability in detection accuracy across different test sets reveals limited model robustness (\textit{i.e.}, demonstrating high sACC). In such cases, providing only binary classification results poses significant risks in practical applications. In contrast, we frame face forgery analysis as a VQA task, requiring forgery analysis prior to judgment, which improves generalization and clarifies the decision-making process. To address fuzzy boundaries between real and forged faces, we integrate hypothetical prompts and MIDS for a comprehensive comparison of answers under different hypotheses, enhancing robustness. Overall, our method significantly improves accuracy and robustness while clarifying the model’s decision-making, enabling more effective application in real-world scenarios.

\renewcommand{\arraystretch}{1.0}
\begin{table}[t]
\centering
\resizebox{.95\columnwidth}{!}{
\begin{tabular}{ccccccc}
\toprule
\textbf{ID} & \textbf{Mask Result} & \textbf{Unfreeze} & \textbf{ACC}$\uparrow$ & \textbf{AUC}$\uparrow$ & \textbf{sACC}$\downarrow$ \\
\midrule
1  &\scalebox{1.2}{-}          &\scalebox{1.2}{-}          & 78.4 & 88.3 & 15.1 \\
2  &\scalebox{1.2}{-}          &\scalebox{1.2}{\checkmark} & 78.2 & 89.4 & 15.3\\
3  &\scalebox{1.2}{\checkmark} &\scalebox{1.2}{-}          & 81.3 & 91.3 & 10.4 \\
4  &\scalebox{1.2}{\checkmark} &\scalebox{1.2}{\checkmark} & \textbf{86.5} & \textbf{94.4} & \textbf{10.0}\\
\bottomrule
\end{tabular}
}
%}
\caption{\small MIDS design choice ablations (\%). We compare the overall generalization performance across the entire OW-FFA-Bench and model robustness across all test sets.}
\label{mids-ablation}
\end{table}

\subsection{Ablation Study}

In this section, we first study the effectiveness of our proposed FFA-VQA dataset. Then, we conduct ablation studies to assess MIDS and its specific design choices.

\begin{figure}[t]
\centering
\includegraphics[width=.98\columnwidth]{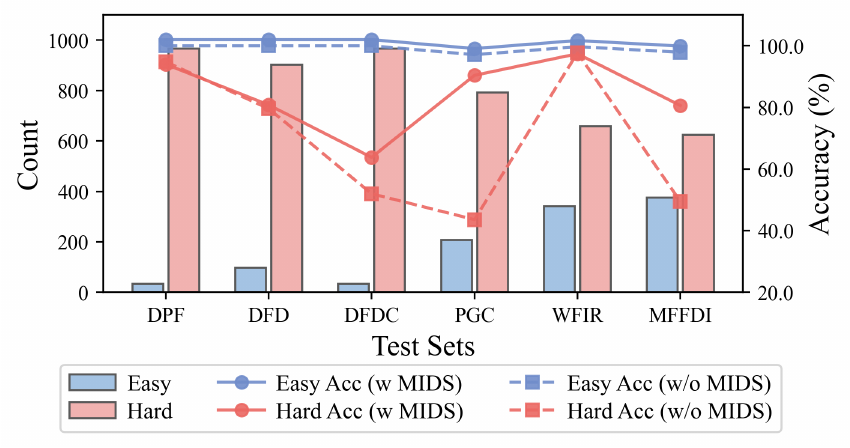}
\caption{The distribution of \textit{'Easy'} and \textit{'Hard'} images as determined by the fine-tuned MLLM and the accuracy improvements on \textit{'Hard'} images achieved by MIDS. On \textit{'Easy'} images, performance is consistent both with and without MIDS.}
\label{fig:easy_hard}
\end{figure}

\boldparagraph{Ablation Study on FFA-VQA}
\label{subsec: ab-ffa-vqa}
We study the effectiveness of the FFA-CoT and the impact of high-quality image descriptions and forgery reasoning textual data on various MLLMs. Specifically, we utilize advanced and open-source MLLMs, including LLaVA-phi-3-mini \cite{phi3} and LLaVA-mistral-7B \cite{llava1.5}. We fine-tune these MLLMs on FFA-VQA containing only \textit{'Analysis results'} (R), \textit{'Image description + Analysis results'} (DR), \textit{'Forgery reasoning + Analysis results'} (FR) and the complete FFA-CoT (All) respectively. We evaluate the fine-tuned MLLMs on the in-domain test set (MA) and the cross-domain test set (OW-FFA-Bench). Fig.~\ref{dataset_quality} presents the detection accuracy (\%) of these MLLMs across various test sets. Relying solely on \textit{'Analysis results'} leads to overfitting and limited generalization. With the inclusion of image descriptions and forgery reasoning, the task evolves into a VQA task that necessitates deeper analysis and reasoning, enhancing generalization on cross-domain test sets. In addition, employing the complete FFA-CoT further improves generalization.

\boldparagraph{Ablation Study on MIDS} 
We first explore the effectiveness of MIDS. Fig.~\ref{fig:easy_hard} presents the distribution of \textit{'Easy'} and \textit{'Hard'} images determined by the MLLM in different test sets of OW-FFA-Bench, reflecting the difficulty of each test set to some extent. After incorporating MIDS, the accuracy on \textit{'Hard'} images significantly improves, effectively enhancing model robustness. 
For \textit{'Easy'} images, accuracy remains high, demonstrating that our method effectively distinguishes between easy and difficult samples without any threshold settings. We then ablate several design choices of MIDS in Tab.~\ref{mids-ablation}. (1) \textit{Mask 'Analysis result' of answers}. Before inputting the answer $\bX_a$ into MIDS, we experiment with leaving the \textit{'Analysis Result'} unmasked, resulting in 78.2\% ACC, which is 8.3\% lower than the masked scenario. We hypothesize that MIDS may be inclined to take shortcuts by focusing on the visible final result. Masking the final result compels the model to infer the relationship between the image and the analysis process. (2) \textit{Unfreeze the last two layers of CLIP-ViT}. We unfreeze the last two layers of visual encoder of MIDS and observe further improvements in generalization. However, without masking, there is no improvement in ACC and model robustness declines, indicating that the CLIP-ViT's ability to perceive face authenticity effectively improves only with attention to enriched textual features from the forgery analysis process.

\subsection{Qualitative Study}
\boldparagraph{Comparison with Advanced MLLMs in Real-World Scenarios}
Fig.~\ref{fig:answer_visual} illustrates an application scenario in public information security. We compare the responses of FFAA with other advanced MLLMs, including LLaVA-Llama-8B \cite{llava} and GPT-4o \cite{gpt4o}. LLaVA demonstrates discrimination capabilities but lacks substantial evidential support. GPT-4o exhibits comprehensive knowledge in face forgery analysis, employing multi-perspective assessments but often avoids definitive judgments, likely due to the inherent ambiguity of forgery analysis and safety policies of GPT. FFAA performs a multi-faceted analysis of facial features, extracting detailed forgery clues, assessing authenticity, and providing a match score and difficulty type as indicators of response reliability.
% These advanced MLLMs exhibit strong image understanding capabilities but often struggle to determine face authenticity, sometimes avoiding definitive conclusions. In contrast, our method not only understands the image but also conducts a reasoned analysis of authenticity from various perspectives.

\begin{figure}[t]
\centering
\includegraphics[width=.95\columnwidth]{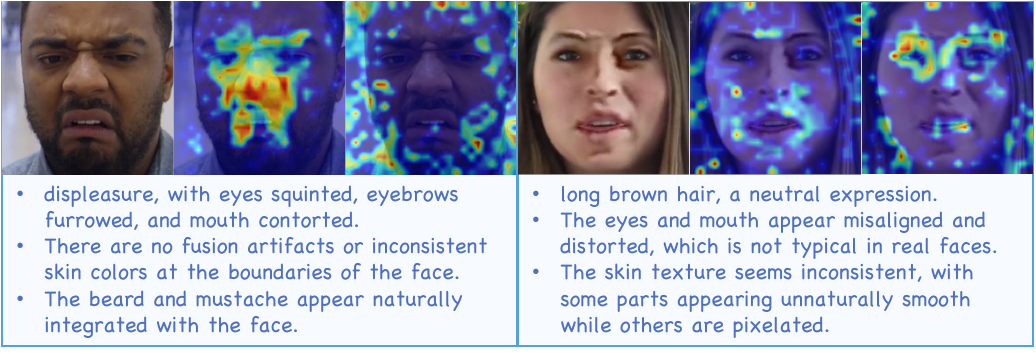}
\caption{\small Attention heatmap visualization. The first row displays two sets of (image, heatmaps for $\bF_{vl}$ and $\bF_{vg}$), while the second row presents some key information from FFAA's answer.}
\label{fig:heatmap}
\end{figure}

\begin{figure}[t]
\centering
\includegraphics[width=.95\columnwidth]{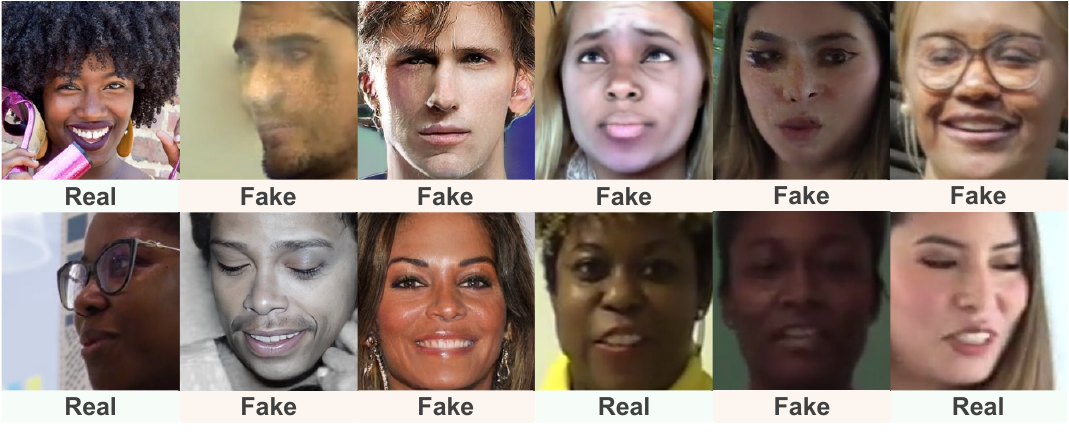}
\caption{\small Easy and hard samples visualization. The first and second rows respectively display the \textit{'Easy'} samples and \textit{'Hard'} samples as determined by FFAA on OW-FFA-Bench.}
\label{fig:easy-hard-samples}
\end{figure}

\boldparagraph{Heatmap Visualization}
To verify whether MIDS can learn more face authenticity representation features guided by enriched textual features, we utilize Grad-CAM \cite{gradcam} to visualize the attention heatmap of $\bF_{vl}$ and $\bF_{vg}$. As illustrated in Fig.~\ref{fig:heatmap}, guided by enriched textual features, MIDS considers multiple aspects of the image and focuses on key features related to face authenticity.

\boldparagraph{Visualization of Easy and Hard Samples}
As illustrated in Fig.~\ref{fig:easy-hard-samples}, the fake faces in the \textit{'Easy'} samples usually have obvious signs of forgery while real faces have higher image quality. Conversely, \textit{'Hard'} samples often exhibit varying image quality and more convincing forgeries, posing significant challenges in determining face authenticity.

\section{Conclusion}
\label{sec:conclusion}

In this paper, we introduce a novel OW-FFA-VQA task aimed at understanding the decision-making process of face forgery analysis models and establish a corresponding benchmark. We first leverage GPT4-assisted analysis generation to create the FFA-VQA dataset, encompassing diverse real and forged face images with essential descriptions and forgery reasoning. Based on this dataset, we introduce FFAA: Face Forgery Analysis Assistant, which consists of a fine-tuned MLLM and MIDS. Our experiments demonstrate that FFAA not only provides user-friendly and explainable results but also significantly enhances generalization, accuracy and robustness compared to previous methods.

\boldparagraph{Limitations} Our method prioritizes detailed analysis and model decision transparency, leading to longer inference times and limited suitability for large-scale forgery screening. Additionally, it is restricted to face images, whereas forgery can involve multiple modalities.

\boldparagraph{Future Works} We will work on reducing inference time while improving accuracy and explore explainable face forgery analysis methods with multi-modal inputs using Multimodal Large Language Models.

% \clearpage
{
    \small
    \bibliographystyle{ieeenat_fullname}
    \bibliography{main}
}

\clearpage
\setcounter{page}{1}
\maketitlesupplementary

% \section{Rationale}
% \label{sec:rationale}
% % 
% Having the supplementary compiled together with the main paper means that:
% % 
% \begin{itemize}
% \item The supplementary can back-reference sections of the main paper, for example, we can refer to \cref{sec:intro};
% \item The main paper can forward reference sub-sections within the supplementary explicitly (e.g. referring to a particular experiment); 
% \item When submitted to arXiv, the supplementary will already included at the end of the paper.
% \end{itemize}
% % 
% To split the supplementary pages from the main paper, you can use \href{https://support.apple.com/en-ca/guide/preview/prvw11793/mac#:~:text=Delete%20a%20page%20from%20a,or%20choose%20Edit%20%3E%20Delete).}{Preview (on macOS)}, \href{https://www.adobe.com/acrobat/how-to/delete-pages-from-pdf.html#:~:text=Choose%20%E2%80%9CTools%E2%80%9D%20%3E%20%E2%80%9COrganize,or%20pages%20from%20the%20file.}{Adobe Acrobat} (on all OSs), as well as \href{https://superuser.com/questions/517986/is-it-possible-to-delete-some-pages-of-a-pdf-document}{command line tools}.

\section{Dataset Construction}

\begin{algorithm}
\small
\caption{FFA-VQA Dataset Construction}
\label{alg:generate_dataset}
\begin{algorithmic}[1]
\Require source dataset $\mathcal{M}$, question prompt set $\mathcal{O}$, queried images set $\mathcal{Q}$, system prompt $S$, threshold $\delta$, budget $N$
\State $(\mathcal{I}, \mathcal{R}) \gets \text{Decompose}(\mathcal{M})$
\State Remaining images set $\mathcal{T} \gets \mathcal{I} \setminus \mathcal{Q}$
\State $k \gets \min(N, |\mathcal{T}|)$
\State Candidate set $\mathcal{C} \gets \emptyset$
\Repeat
    \State Randomly select a question prompt $x_q \in \mathcal{O}$
    \State Sample an image $I$ from dataset $\mathcal{T}$
    \State $(L, F) \gets \text{ParseImage}(I)$ \Comment{$L$: label, $F$: forgery type}
    \State $x_f \gets \text{GetForgerySpecificPrompt}(L, F)$
    \If{$F$ is \texttt{'IE'} or \texttt{'FAM'}}
    \State $(R, x_r) \gets \text{GetReferenceAndPrompt}(I, \mathcal{R})$
    \State $x \gets (I, x_f, R, x_r, x_q)$
    \Else
    \State $R \gets \text{None}$
    \State $x \gets (I, x_f, x_q)$
    \EndIf
    \State Get answer $A \gets \text{GPT4o}(S, x)$
    \State $\mathcal{Q} \gets \mathcal{Q} \cup \{I\}, \; \mathcal{T} \gets \mathcal{T} \setminus \{I\}, \; k \gets k - 1$
    \If{\text{$A$ does not follow FFA-CoT format}}
    \State \textbf{continue}
    \EndIf
    \State $(L_a, F_a, P_a) \gets \text{ParseAnswer}(A)$
    \If{$(L_a = L)$ and $(F_a = F)$ and $(P_a \geq \delta)$}
    \State $\mathcal{C} \gets \mathcal{C} \cup \{(I, R, L, F, x_q, A)\}$
    \Else
    \State \textbf{continue}
    \EndIf
\Until{$k\leq0$}
\State $\mathcal{C}' \gets \emptyset$
\For{\text{each } $(I, R, L, F, x_q, A) \in \mathcal{C}$}
    \If{\text{Expert approves $A$ based on $(I, R, L, F)$}}
        \State $\mathcal{C}' \gets \mathcal{C}' \cup \{(I, x_q, A)\}$
    \EndIf
\EndFor
\State \textbf{Return} Qualified VQA dataset $\mathcal{C}'$
\end{algorithmic}
\end{algorithm}

\begin{table}[t]
\centering
\resizebox{.95\columnwidth}{!}{
\begin{tabular}{lll}
\toprule
Type  & Source Datasets  & Samples \\
\midrule
Real  & \begin{tabular}[l]{@{}l@{}}FF++\cite{ff++/xception} (Original)\\ Celeb-DF-v2\cite{celeb} (Celeb-real, Youtube-real)\\ DFFD\cite{dffd} (FFHQ)\end{tabular} & 27K  \\
\midrule
\begin{tabular}[l]{@{}l@{}}Identity\\ exchange\end{tabular} & \begin{tabular}[l]{@{}l@{}}FF++\cite{ff++/xception} (Deepfakes, FaceSwap)\\ Celeb-DF-v2\cite{celeb} (Celeb-synthesis)\end{tabular} & 25K  \\
\midrule
\begin{tabular}[l]{@{}l@{}}Facial attribute\\ manipulation\end{tabular} & \begin{tabular}[l]{@{}l@{}}FF++\cite{ff++/xception} (Face2Face, NeuralTexture)\\ DFFD\cite{dffd} (Faceapp)\end{tabular}  & 18K  \\
\midrule
\begin{tabular}[l]{@{}l@{}}Entire face\\ synthesis\end{tabular} & \begin{tabular}[l]{@{}l@{}}DFFD\cite{dffd} (StyleGAN-FFHQ)\\ GanDiffFace\cite{gandiffface}\end{tabular} & 24K \\
\bottomrule
\end{tabular}
}
%}
\caption{Multi-attack dataset.}
\label{tab: multiattack}
\end{table}

\begin{figure}[t]
\centering
\includegraphics[width=.85\columnwidth]{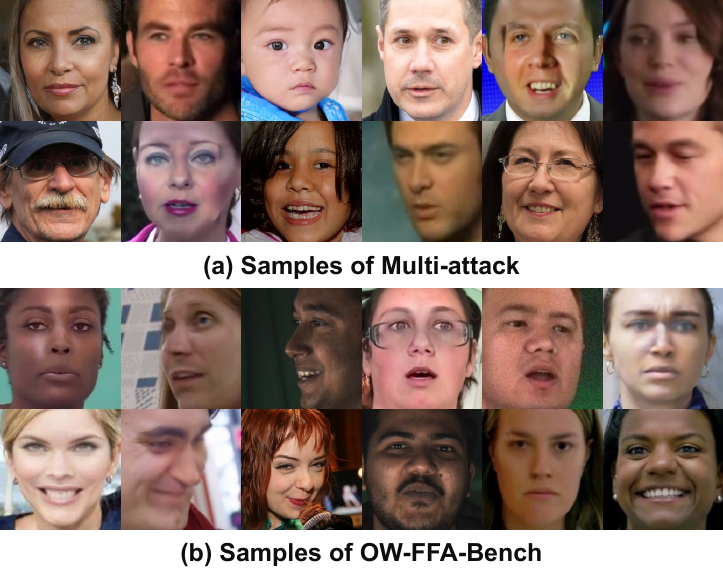}
\caption{Randomly sampled face images.}
\label{fig:data-samples}
\end{figure}

\boldparagraph{Data Source Composition for FFA-VQA} 
We leverage data gathered from the public datasets illustrated in Fig.~\ref{dataset_pipeline}a. Using the dlib library, we extract facial regions from video frames and images, subsequently resizing them uniformly to 224×224. Fig.~\ref{fig:data-samples}a presents examples of these processed face images. The images are classified by forgery type based on their respective sources, as detailed in Tab.~\ref{tab: multiattack}. From this dataset, we randomly select 2,000 images, dividing them equally into a validation set and a test set.
Following the process outlined in Algorithm~\ref{alg:generate_dataset} and Fig.~\ref{dataset_pipeline}b, and using the training set of Multi-attack as the image data source, we construct the FFA-VQA dataset, comprising 20,142 pairs of VQA data distributed as follows: \textit{'Real'} (8,073), \textit{'Facial attribute manipulation'} (4,014), \textit{'Entire face synthesis'} (4,025), and \textit{'Identity exchange'} (4,030). To balance the difficulty and reliability of face forgery analysis, we set the threshold $\delta$ to 0.6. The average number of tokens in the answers is 203.

\boldparagraph{Data Source Composition for OW-FFA-Bench} 
The data sources for OW-FFA-Bench are shown in Tab.~\ref{benchmark}. Similarly, we employ the dlib tool to crop faces from video frames or images and uniformly resize them to 224x224. Finally, to balance testing efficiency and effectiveness, we randomly sample 1,000 images from each dataset according to its data distribution. Fig.~\ref{fig:data-samples}b shows some of the face images from the test sets.

\section{Implementation Details}

\boldparagraph{Fine-tuning Details of the MLLM}
We employ LLaVA-v1.6-mistral-7B \cite{llava1.5} as our backbone and fine-tune it with LoRA (rank=32, alpha=48) on the FFA-VQA dataset containing hypothetial prompts using 2 RTX 3090 GPUs, with a learning rate of 1e-4 and a batch size of 16. All other settings remain the same as in LLaVA.

\boldparagraph{Training Details of MIDS}
We first employ the fine-tuned MLLM to mine historical answer data from the remaining available images in the Multi-attack training set, excluding the \textit{'Easy'} images. Ultimately, we collect 30K $(\bX_v, \bX_a^a, \bX_a^p, \bX_a^n)$ data pairs. We then train MIDS on this dataset for 2 epochs using 2 RTX 3090 GPUs, with the AdamW optimizer, a learning rate of 1e-4, a batch size of 48, and a weight decay of 1e-5. The number of self-attention iterations $M=3$. Data augmentation techniques include \textit{ImageCompression}, \textit{GaussianNoise}, \textit{MotionBlur}, \textit{GaussianBlur}, \textit{FancyPCA} and \textit{HueSaturationValue}.

\boldparagraph{Training Details of Baseline} 
The training details for the methods we trained on the Multi-attack training set are as follows, with all unmentioned details consistent with the code provided by the original authors:
\begin{itemize}
    \item \textit{ResNet-34}: We trained for 20 epochs on 2 RTX 3090 GPUs using the Adam optimizer, with a learning rate of 1e-3, a batch size of 100, and a weight decay of 1e-6. Data augmentation included \textit{HorizontalFlip}.
    \item \textit{EfficientNet-B4}: We trained for 20 epochs on 2 RTX 3090 GPUs using the Adam optimizer, with a learning rate of 1e-3, a batch size of 128, and a weight decay of 1e-6. Data augmentation included \textit{HorizontalFlip}.
    \item \textit{Xception}: We trained for 20 epochs on 2 RTX 3090 GPUs using the Adam optimizer, with a learning rate of 1e-3, a batch size of 112, and a weight decay of 1e-6. Data augmentation included \textit{HorizontalFlip}.
    \item \textit{RECCE}: We trained for 22,500 steps on 2 RTX 3090 GPUs using the Adam optimizer, with a learning rate of 2e-4, a batch size of 32, and a weight decay of 1e-5. 
    \item \textit{DCL}: We trained for 30 epochs on 2 RTX 3090 GPUs using the Adam optimizer, with a learning rate of 2e-4, a batch size of 32, and a weight decay of 1e-5. Additionally, the parameter \texttt{has\_mask} was set to \texttt{false}.
    \item \textit{Exposing}: We trained for 50 epochs on 2 RTX 3090 GPUs using the Adam optimizer, with a learning rate of 1e-4, a batch size of 256, and a weight decay of 1e-6. 
    \item \textit{NPR}: We trained for 98 epochs on 1 RTX 3090 GPU using the Adam optimizer, with a learning rate of 2e-4, a batch size of 32. 
    \item \textit{UIA-ViT}: We trained for 8 epochs on 1 RTX 3090 GPU using the Adam optimizer, with a learning rate of 3e-5, a batch size of 96.
    \item \textit{CLIPping}: We trained for 2 epochs on 1 RTX 3090 GPU using the SGD optimizer, with a learning rate of 2e-3, a batch size of 16.
    \item \textit{F3Net}: We trained for 6 epochs on 1 RTX 3090 GPU using the Adam optimizer, with a learning rate of 2e-4, a batch size of 128.
    \item \textit{Two-Stream}: We trained for 5 epochs on 2 RTX 3090 GPUs using the Adam optimizer, with a learning rate of 2e-4, a batch size of 64, and a weight decay of 1e-6. Data augmentation included \textit{HorizontalFlip}.
    \item \textit{FADD}: We trained for 40 epochs on 1 RTX 3090 GPUs using the Adam optimizer, with a learning rate of 1e-3, a batch size of 64.
\end{itemize}

\section{Answer Analysis}
\begin{figure*}[t]
\centering
\includegraphics[width=.92\textwidth]{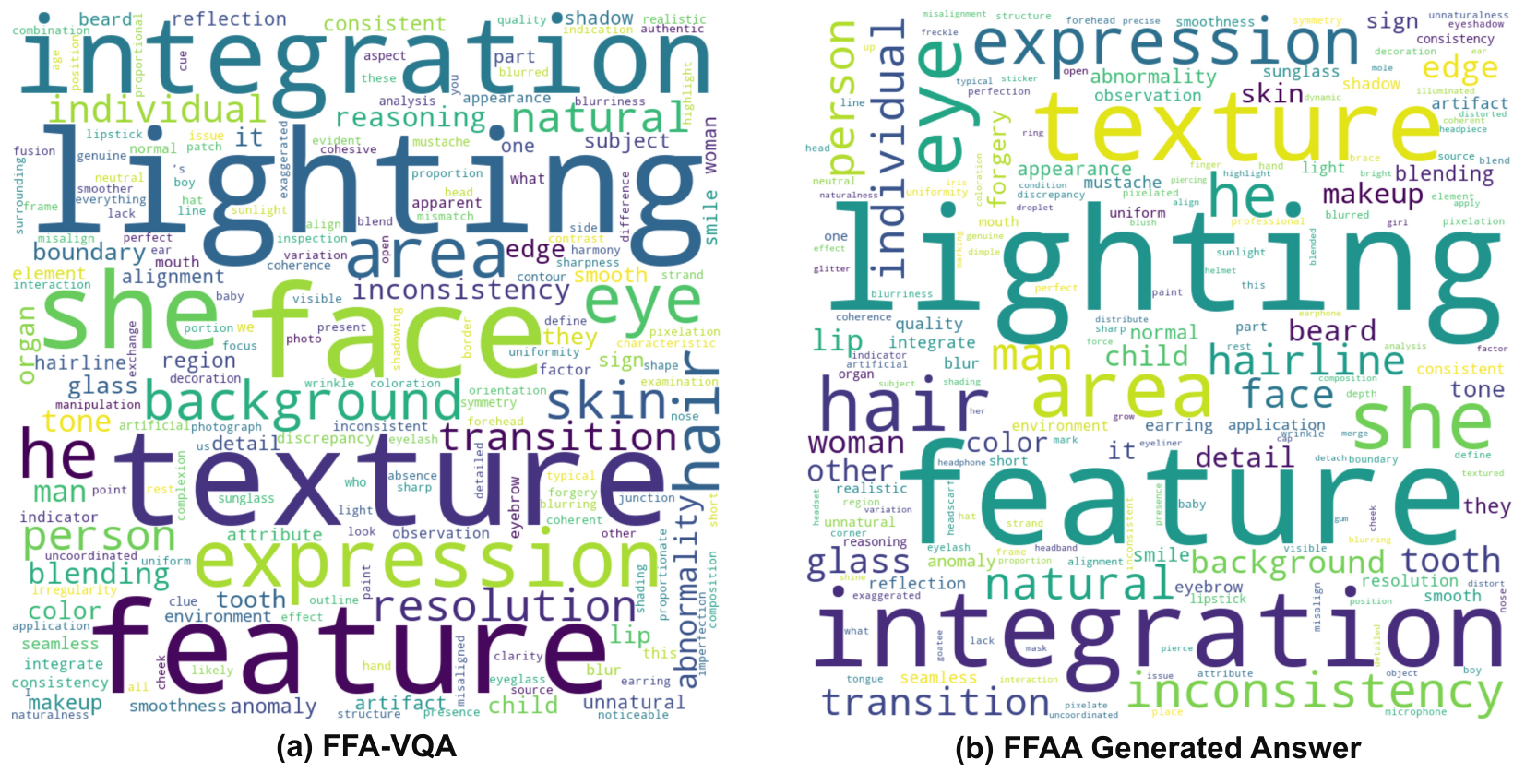}
\caption{\textbf{Analysis of answers generated by GPT-4o and FFAA}.The word cloud distribution of nouns and adjectives from the noun-verb or adjective-verb pairs in the answers. The semantic richness of FFAA's answers is comparable to that of GPT-4o.}
\label{fig:wordcloud}
\end{figure*}

Fig.~\ref{fig:wordcloud} shows the word cloud of nouns and adjectives in the noun-verb and adjective-verb pairs from the answers in FFA-VQA and generated by FFAA.
It is evident that the answers generated by GPT-4o and FFAA are semantically enriched and diverse, encompassing descriptions from multiple aspects. These include descriptions of facial features (\textit{e.g.}, \textit{'eye'}, \textit{'hair'}); forgery-related characteristics (\textit{e.g.}, \textit{'artifact'}, \textit{'sharpness'}); and factors related to image quality and the physical environment (\textit{e.g.}, \textit{'lighting'}, \textit{'resolution'}).

\section{Prompts}
\begin{figure*}[t]
\centering
\includegraphics[width=0.95\textwidth]{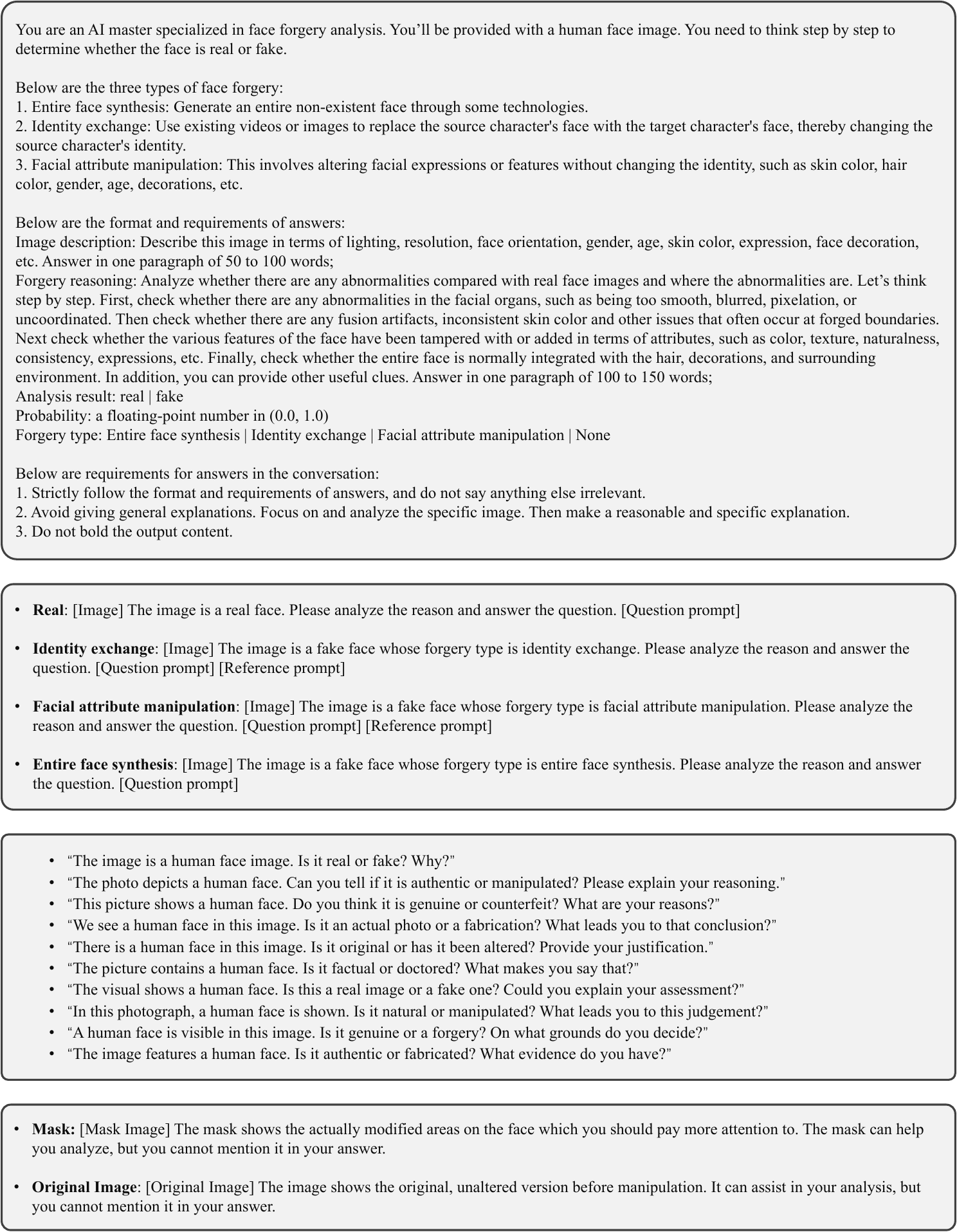}
\caption{From top to bottom: system prompt, forgery-specific prompt, question prompt, reference prompt.}
\label{fig:prompts}
\end{figure*}

The system prompt, forgery-specific prompt, question prompt and reference prompt are shown in Fig.~\ref{fig:prompts}.

\section{Examples}
\begin{figure*}[t]
\centering
\includegraphics[width=.98\textwidth]{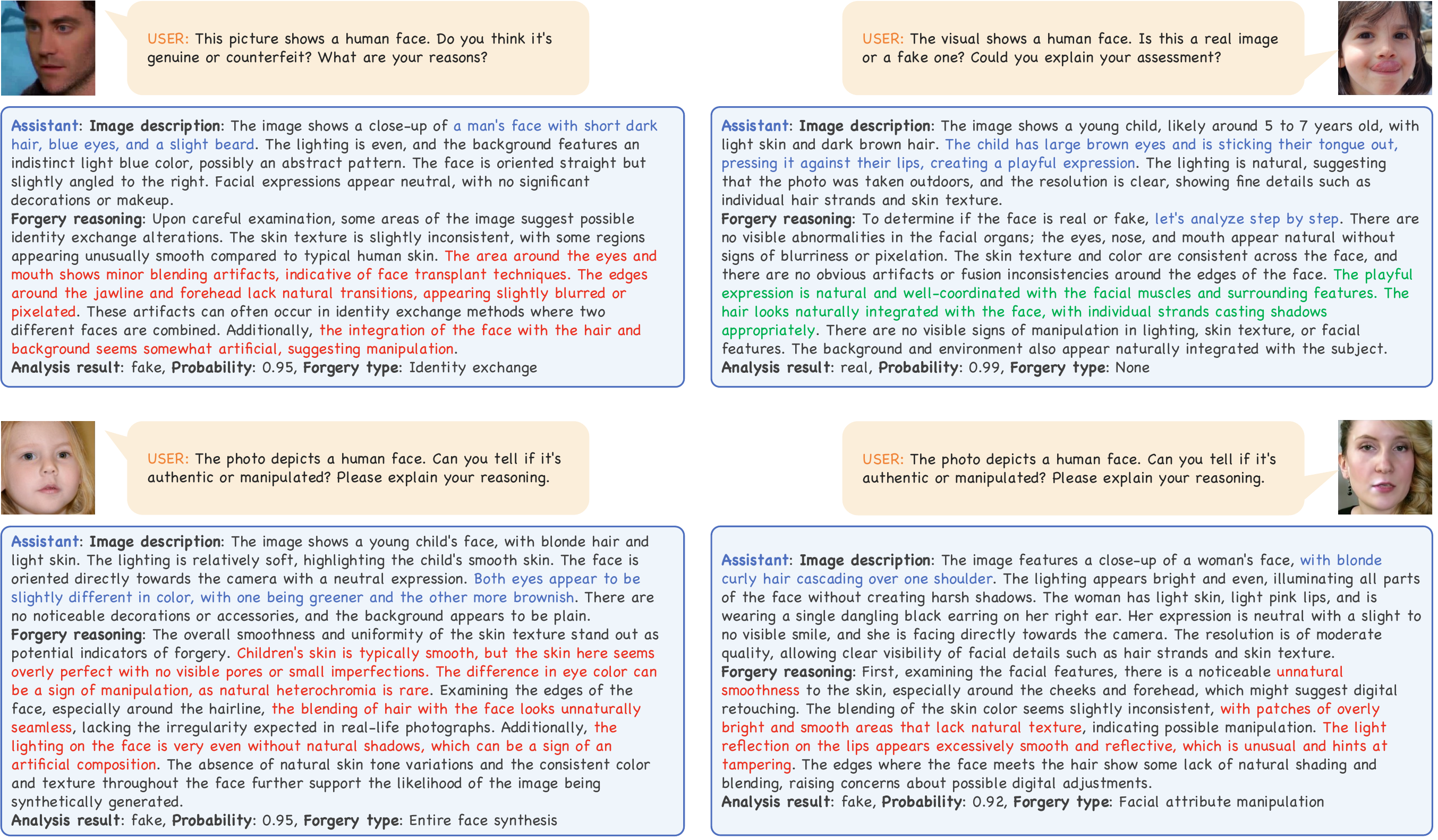}
\caption{Examples of FFA-VQA.}
\label{fig:ffavqa_examples}
\end{figure*}

\boldparagraph{Examples of FFA-VQA dataset}
Fig.~\ref{fig:ffavqa_examples} illustrates examples of VQA pairs for four categories of faces in the FFA-VQA dataset. It can be observed that the image descriptions comprehensively detail facial features, image quality, and surrounding environments. Additionally, the forgery reasoning section provides robust evidence for analysis.

\begin{figure*}[t]
\centering
\includegraphics[width=.95\textwidth]{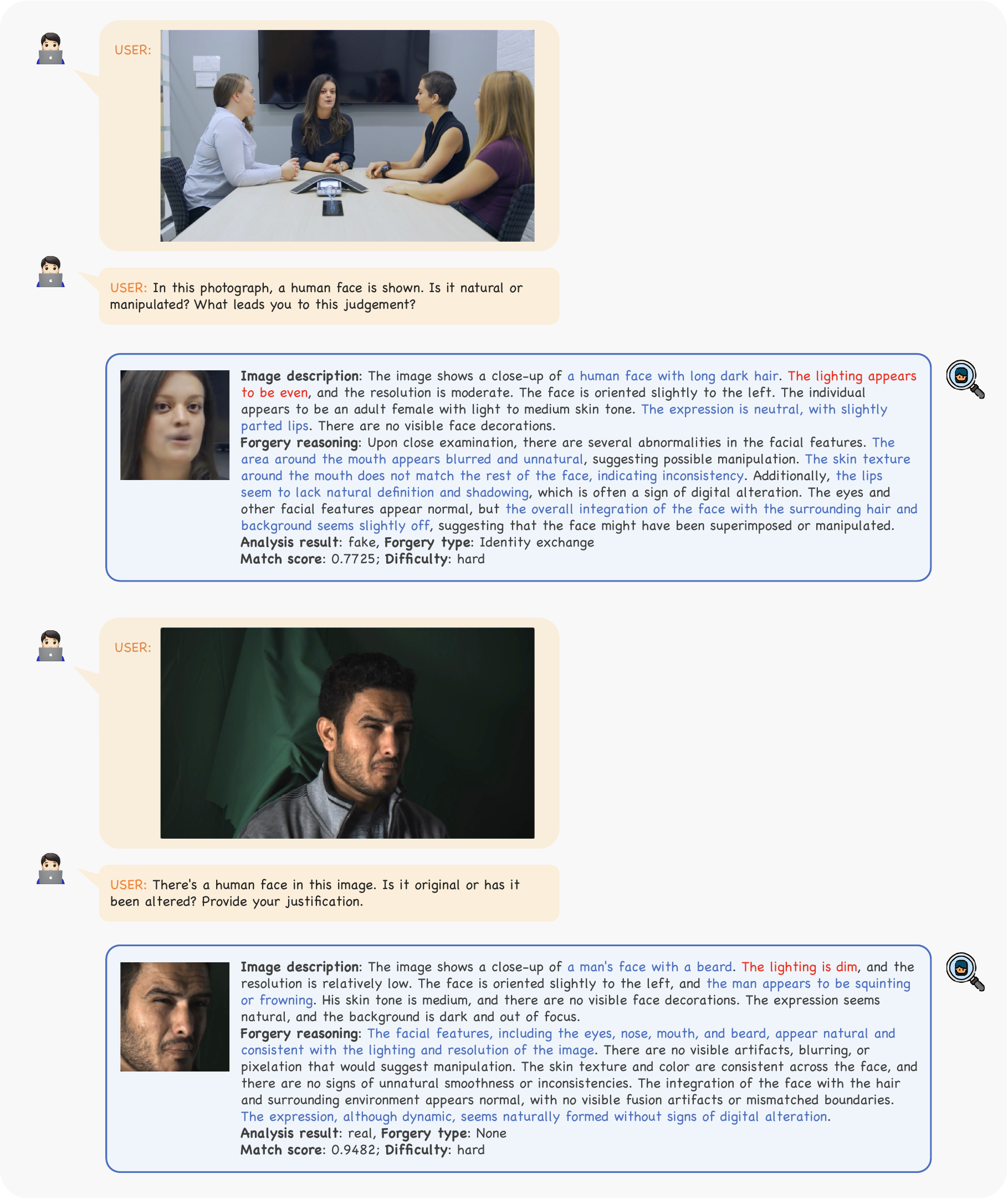}
\caption{\textbf{Dialogue examples of FFAA under different lighting conditions}. The first image, a forged face with the forgery type of identity exchange, is sourced from DeepFakeDetection \cite{deepfakedetection}, while the second image, showing a real face, is sourced from Deeperforensics \cite{deeperforensics}.}
\label{fig:ffaa_example1}
\end{figure*}

\begin{figure*}[t]
\centering
\includegraphics[width=.95\textwidth]{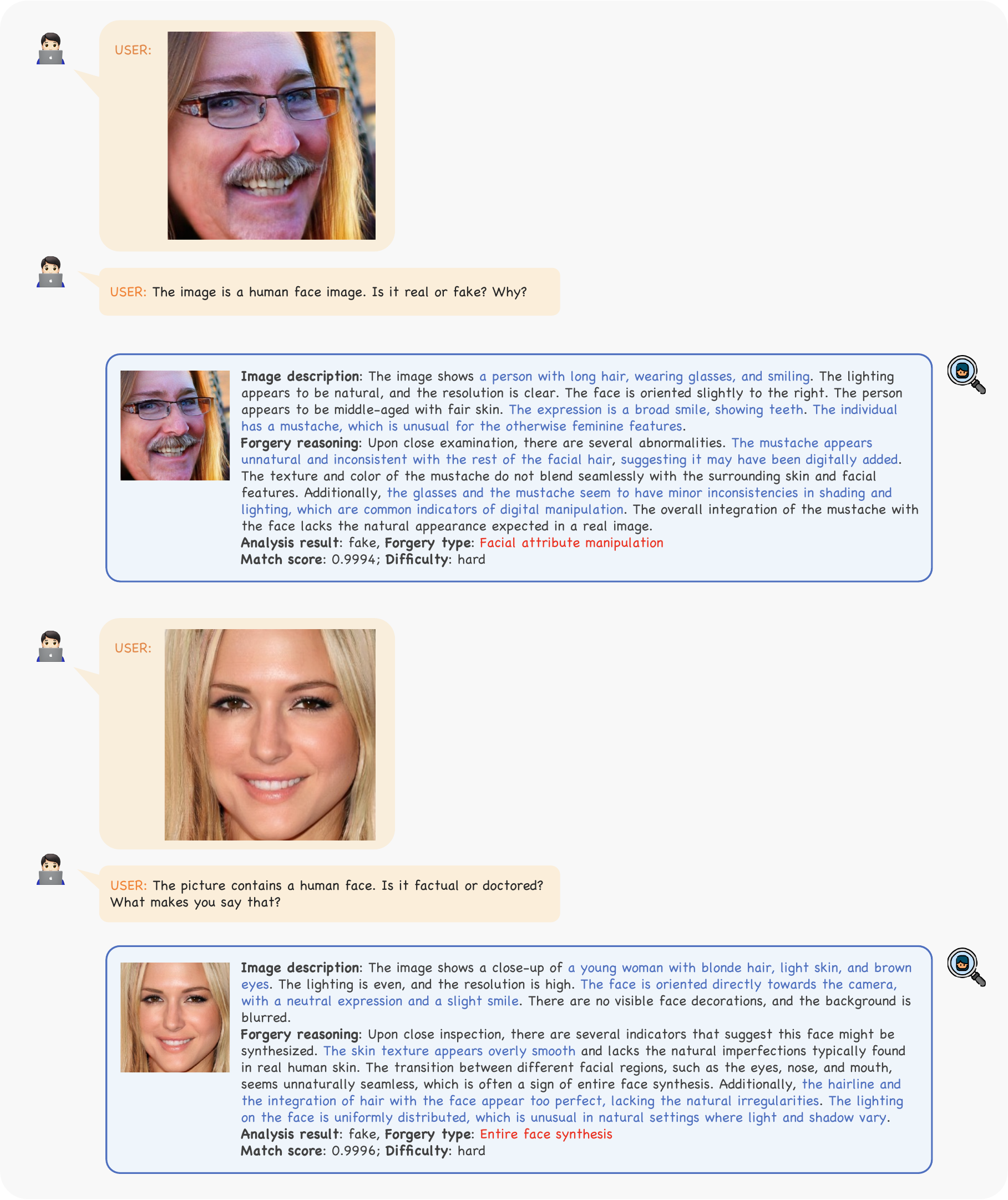}
\caption{\textbf{Dialogue examples of FFAA in scenarios involving malicious photo editing and synthetic faces}. The first image, a forged face with the forgery type of facial attribute manipulation generated using FaceApp \cite{faceapp}, while the second image, showing a synthetic face generated using StyleGAN \cite{stylegan/ffhq}, both are sourced from DFFD \cite{dffd}.}
\label{fig:ffaa_example2}
\end{figure*}

\begin{figure*}[t]
\centering
\includegraphics[width=.95\textwidth]{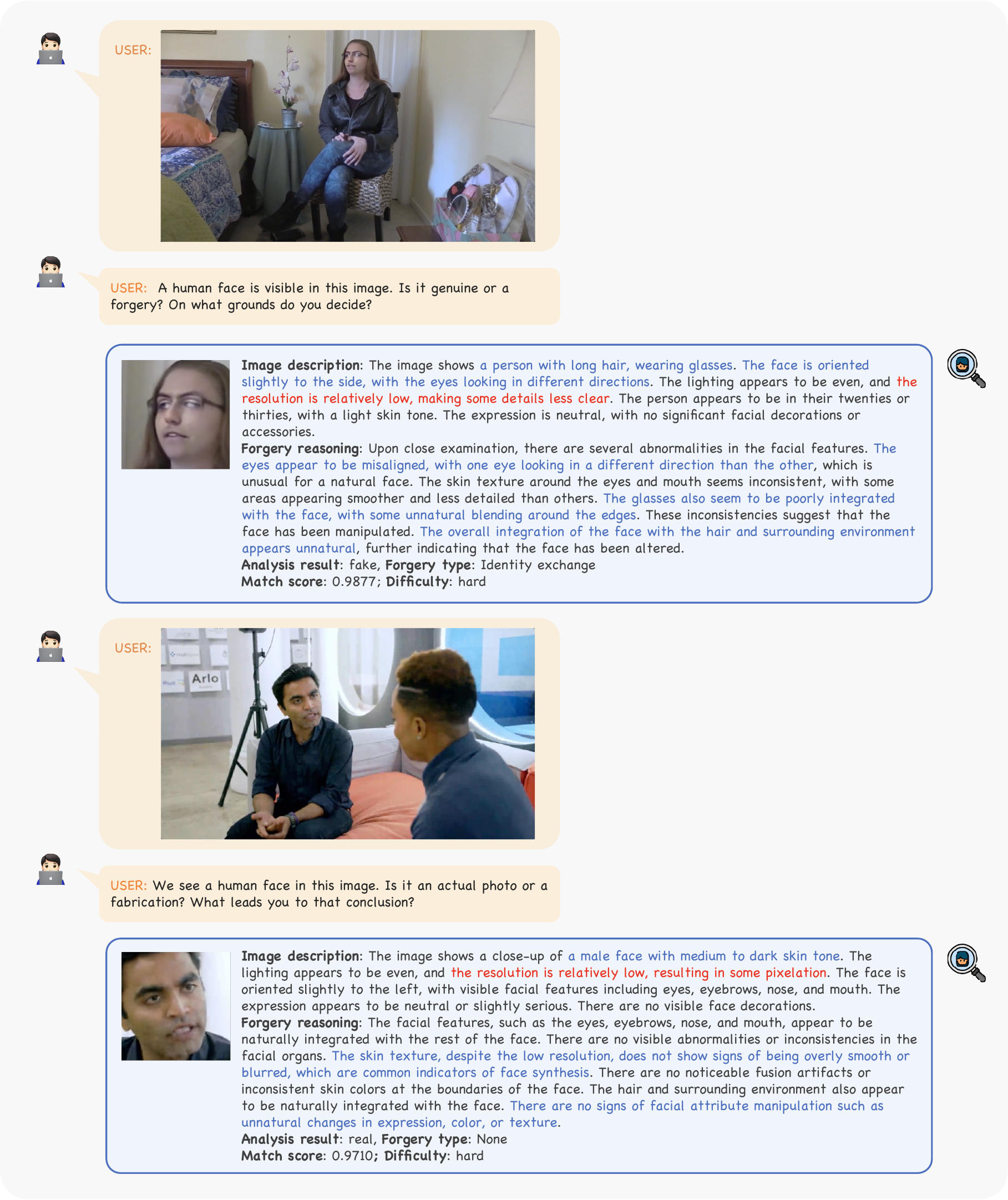}
\caption{\textbf{Dialogue examples of FFAA in low-quality image scenarios}. The first image, a forged face with the forgery type of identity exchange, is sourced from DFDC \cite{dfdc}, while the second image, showing a real face, is sourced from Deeperforensics \cite{deeperforensics}.}
\label{fig:ffaa_example3}
\end{figure*}

\begin{figure*}[t]
\centering
\includegraphics[width=.95\textwidth]{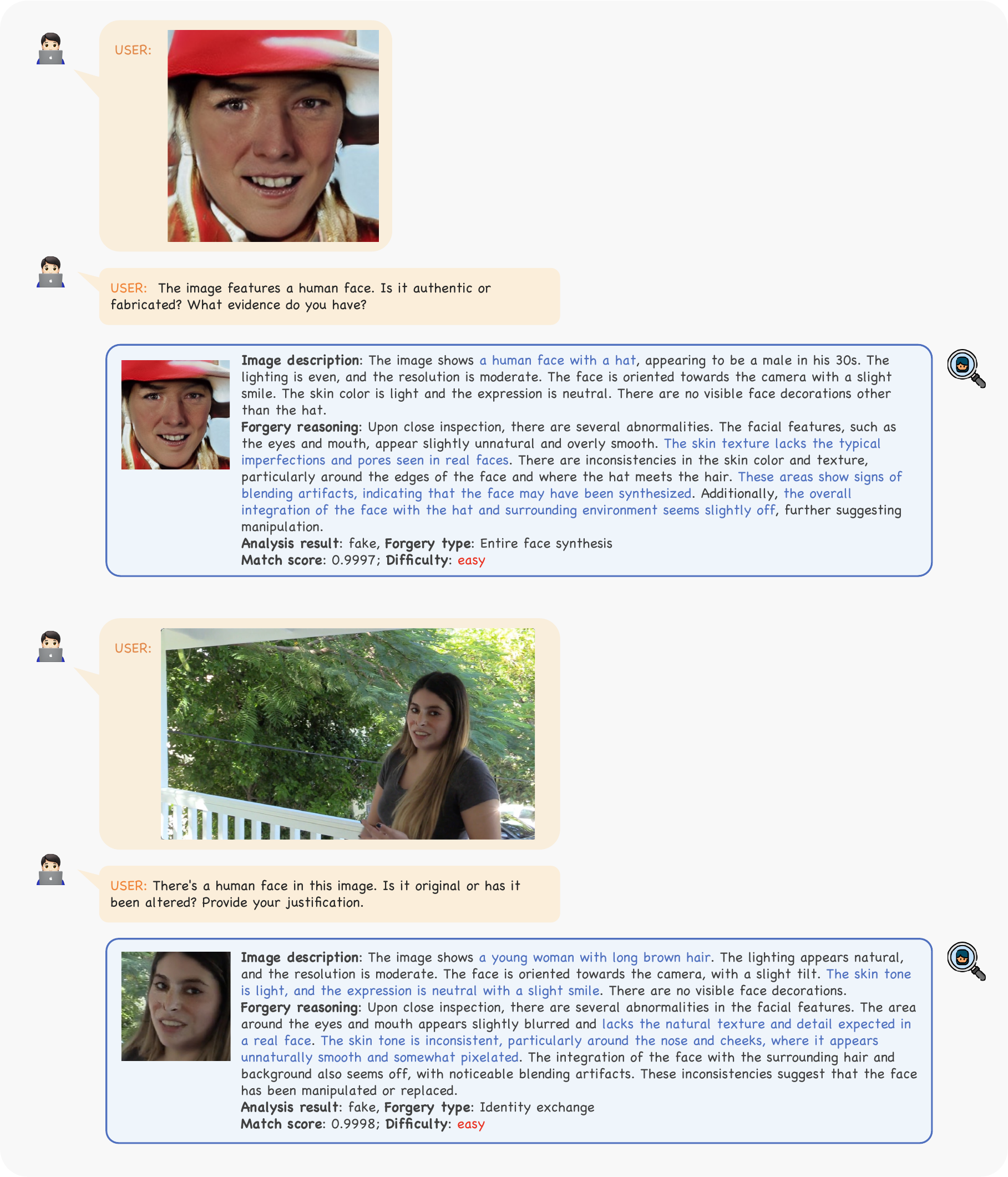}
\caption{\textbf{Dialogue examples of FFAA in scenarios involving obvious forgeries}. The first image, a synthetic face generated using StyleGAN \cite{stylegan/ffhq}, is sourced from DFFD \cite{dffd}, while the second image, a forged face with the forgery type of identity exchange, is sourced from DFDC \cite{dfdc}.}
\label{fig:ffaa_example4}
\end{figure*}

\begin{figure*}[t]
\centering
\includegraphics[width=.95\textwidth]{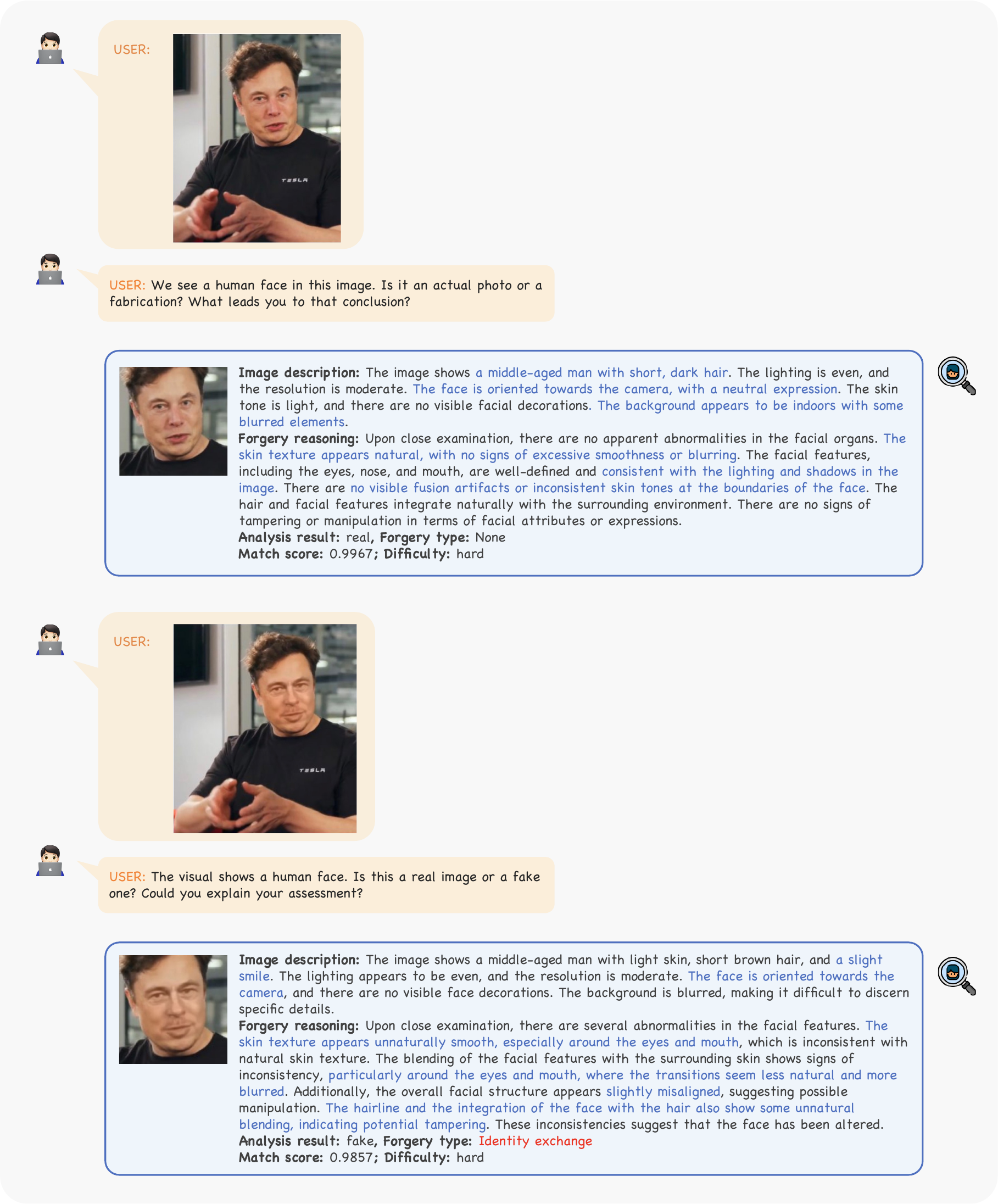}
\caption{\textbf{Dialogue examples of FFAA in real-world scenarios involving celebrity face-swapping}. The first image features Elon Musk's face sourced from the internet, while the second image is a forged face replaced using an unknown forgery technique by another person.}
\label{fig:ffaa_example5}
\end{figure*}

\begin{figure*}[t]
\centering
\includegraphics[width=.95\textwidth]{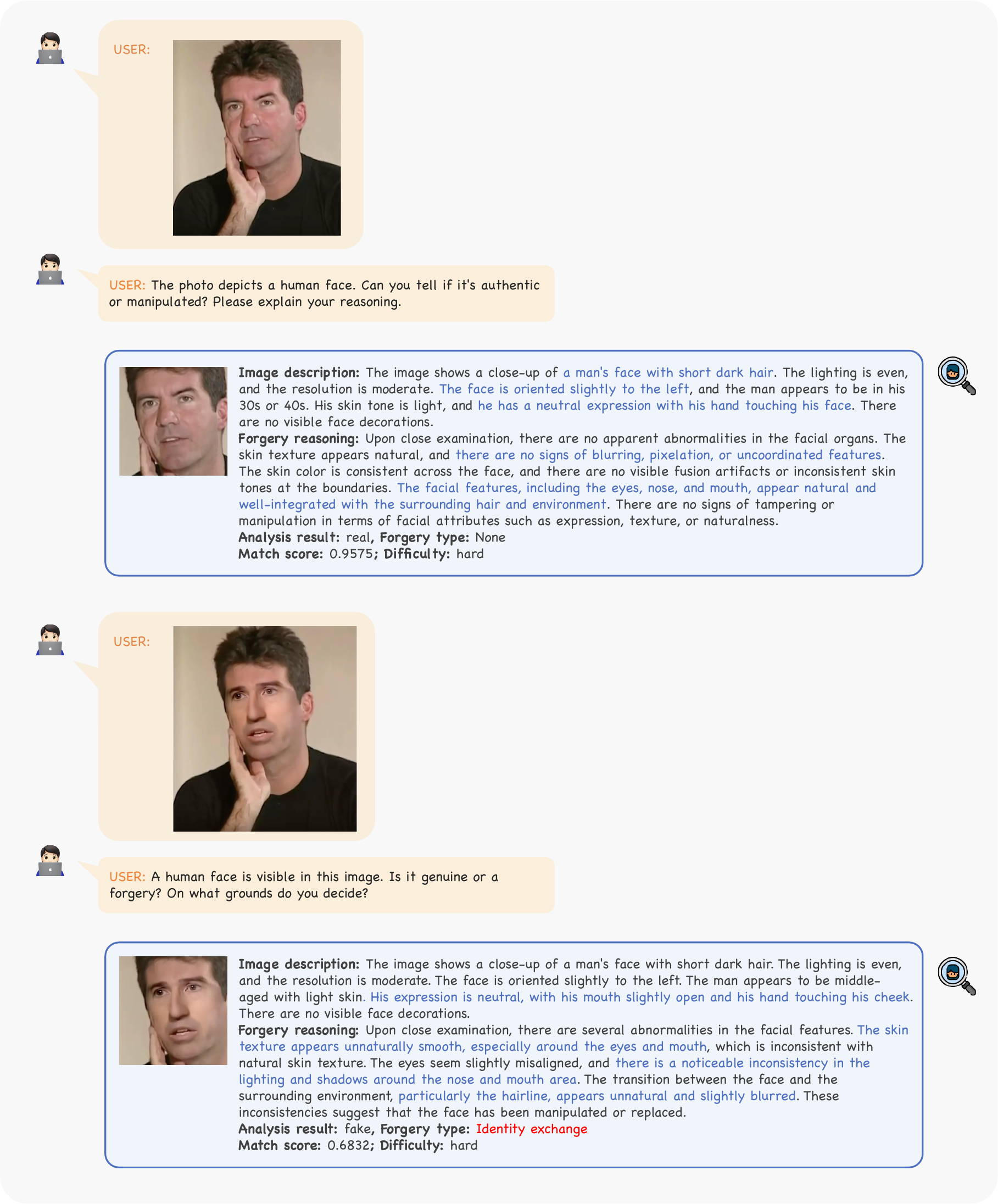}
\caption{\textbf{Dialogue examples of FFAA in real-world scenarios involving celebrity face-swapping}. The first image features a real face sourced from the internet, while the second image is a forged face replaced using an unknown forgery technique by another person.}
\label{fig:ffaa_example6}
\end{figure*}

\begin{figure*}[t]
\centering
\includegraphics[width=.95\textwidth]{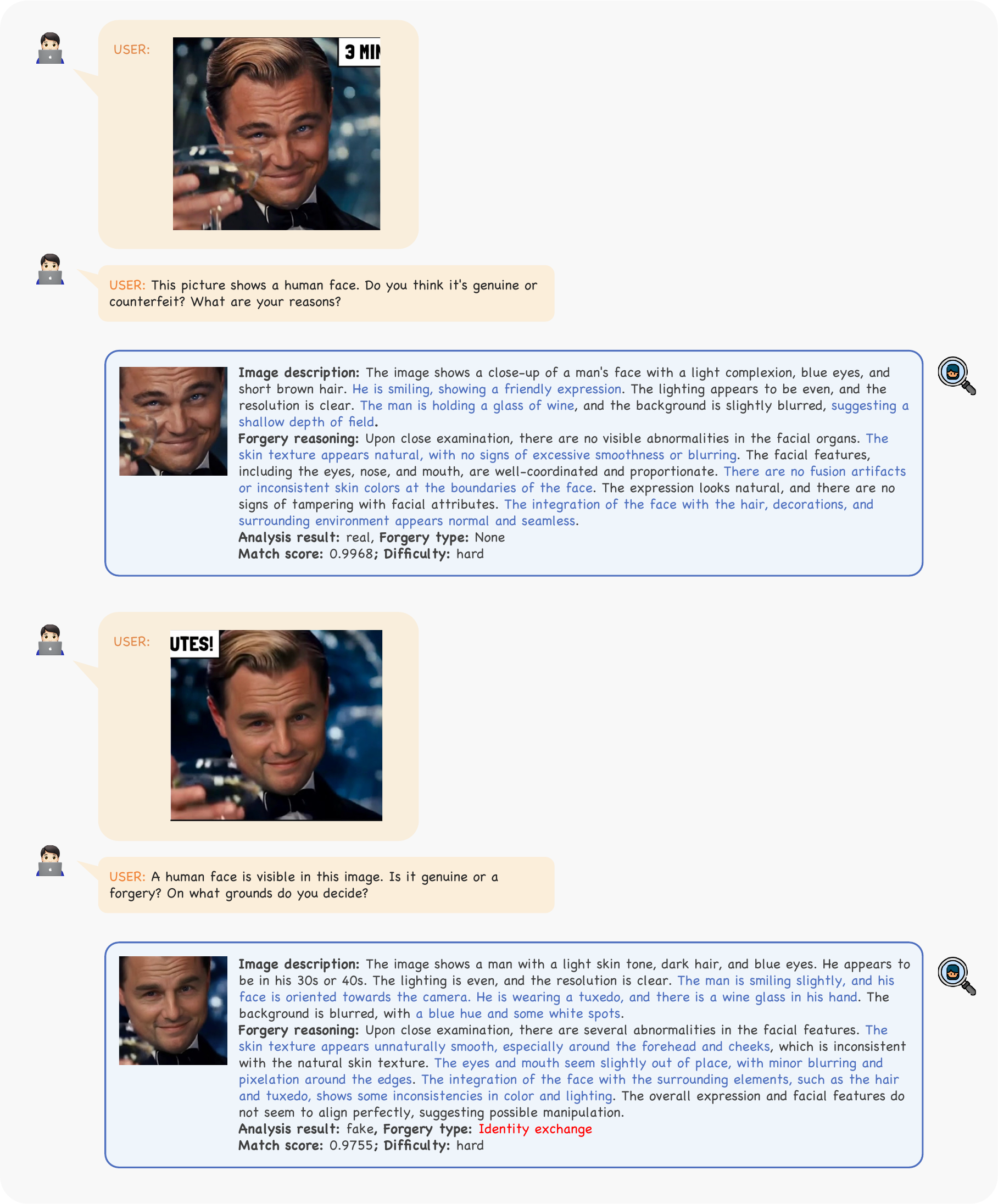}
\caption{\textbf{Dialogue examples of FFAA in real-world scenarios involving celebrity face-swapping}. The first image is a facial screenshot of Leonardo DiCaprio from a movie, sourced from the internet, while the second image is a forged face replaced using an unknown forgery technique by another person.}
\label{fig:ffaa_example7}
\end{figure*}

\begin{figure*}[t]
\centering
\includegraphics[width=.95\textwidth]{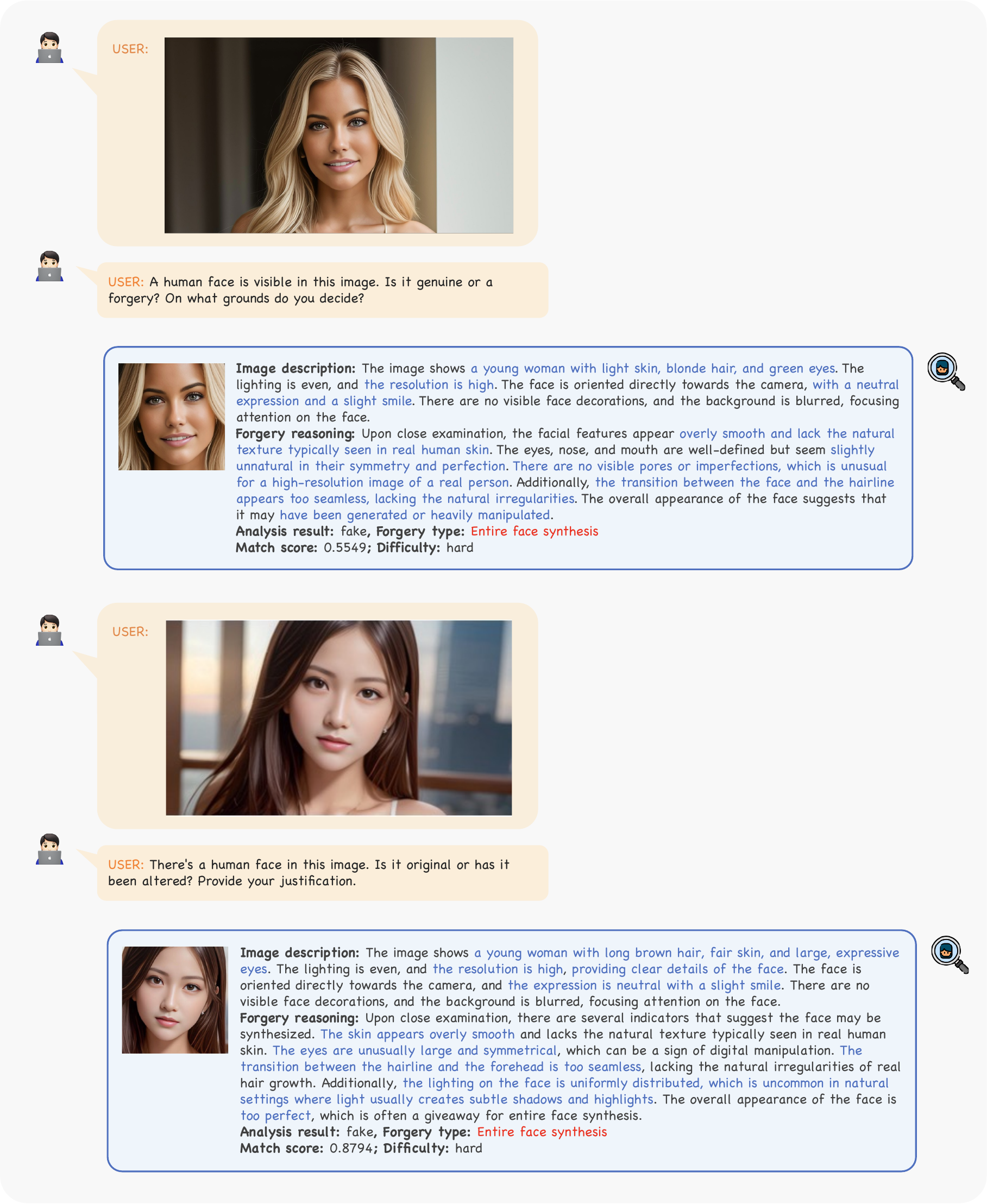}
\caption{\textbf{Dialogue examples of FFAA in real-world scenarios involving AIGC-generated faces}. Both images are high-fidelity depictions of beauty, sourced from the internet and created using unknown generation techniques.}
\label{fig:ffaa_example8}
\end{figure*}

\boldparagraph{Dialogue Examples of FFAA}
Figures \ref{fig:ffaa_example1} to \ref{fig:ffaa_example4} present FFAA dialogue examples from the OW-FFA-Bench and Multi-attack test set, covering scenarios such as varied lighting, image editing, generated images, low resolution, and evident forgery traces.
Figures \ref{fig:ffaa_example5} to \ref{fig:ffaa_example8} highlight FFAA's performance against real-world unknown forgery techniques, including celebrity face swaps and high-fidelity AIGC-generated faces. These examples demonstrate FFAA's ability to provide strong evidence and accurately identify forgery types, even when faced with unknown deepfake technologies.

\begin{figure*}[t]
\centering
\includegraphics[width=1.0\textwidth]{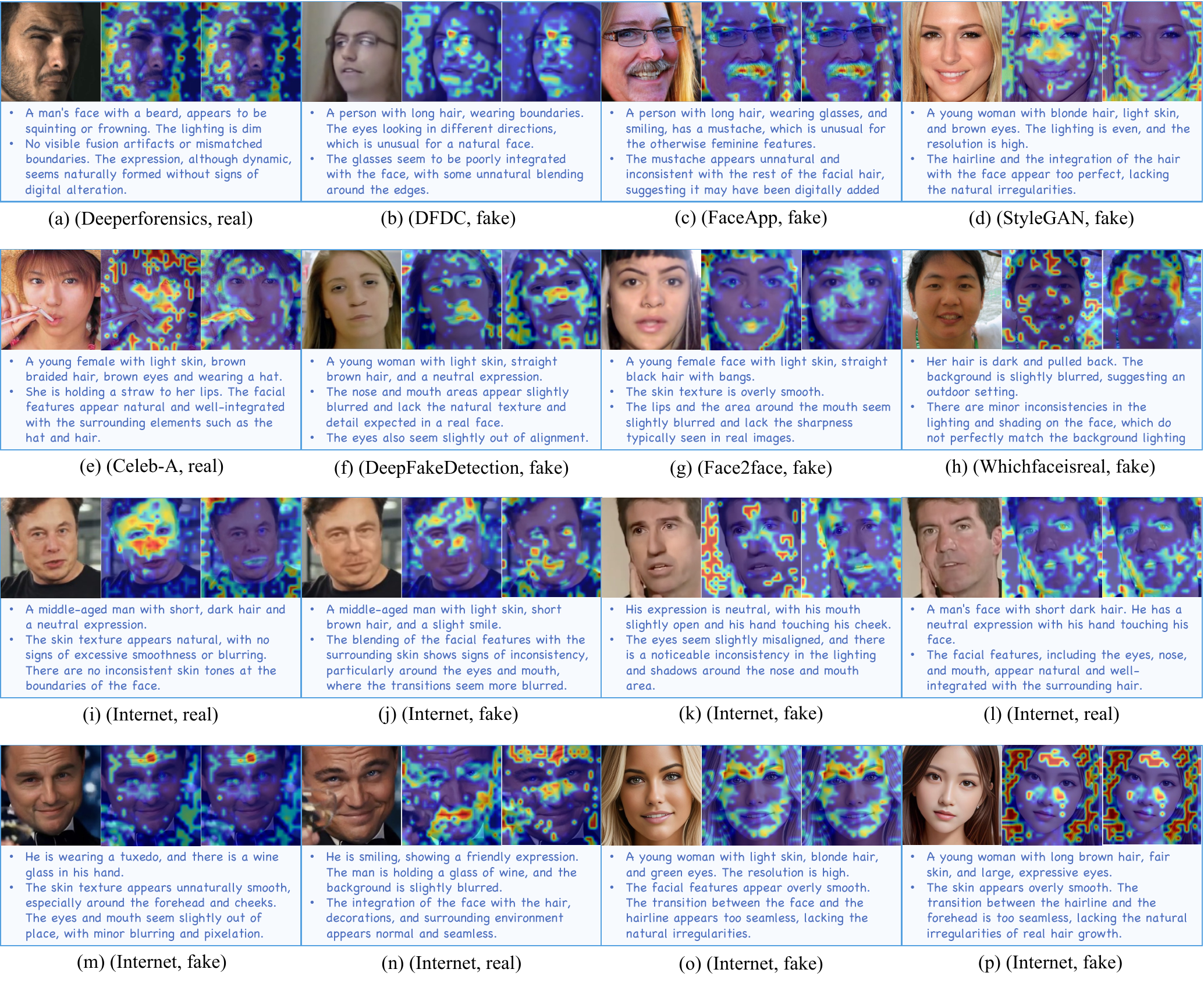}
\caption{\textbf{More heatmaps visualization}. We present sixteen face images from different sources, along with the heatmap visualizations of MIDS's $\bF_{vl}$ and $\bF_{vg}$, and several key information from FFAA's answers. It can be observed that $\bF_{vl}$ tends to focus on specific facial features, the surrounding environment, and more apparent local forgery marks or authenticity clues, whereas $\bF_{vg}$ emphasizes more abstract features (\textit{e.g.}, \textit{'smoothness'}, \textit{'lighting'}, \textit{'shadows'}, \textit{'integration'}). In some cases, such as examples (a), (b), (c), (l), (m), (o) and (p), $\bF_{vl}$ and $\bF_{vg}$ may rely on the same features as clues for assessing the authenticity of the face.}
\label{fig:more_heatmaps}
\end{figure*}

\boldparagraph{More Heatmaps Visualization}
Fig.~\ref{fig:more_heatmaps} presents additional heatmap visualization results, further illustrating how MIDS, guided by rich semantic analysis, effectively captures detailed image features and diverse, deeper forgery characteristics.

\end{document}